\documentclass[10pt,twocolumn,twoside]{IEEEtran}

\usepackage{amssymb}
\usepackage{amsthm}

\usepackage[pdftex]{graphicx}
\ifCLASSINFOpdf
\else
\fi

\usepackage{subfigure}
\usepackage[cmex10]{amsmath}

\usepackage{algorithm}
\usepackage{algorithmic} 
\usepackage{array}
\usepackage{mdwmath}
\usepackage{mdwtab}
\usepackage{url}

\begin{document}
%
\title{Learning Sampling Distributions for\\ Efficient Object Detection}

\author{Yanwei~Pang,~\IEEEmembership{Senior~Member,~IEEE}, Jiale~Cao, and~Xuelong~Li,~\IEEEmembership{Fellow,~IEEE}
       
%
\thanks{Y. Pang and J. Cao are with the School of Electronic Information Engineering, Tianjin University, Tianjin 300072, China. e-mails: \{pyw,connor\}@tju.edu.cn}

\thanks{X. Li is with the Center for OPTical IMagery Analysis and Learning (OPTIMAL), State Key Laboratory of Transient Optics and Photonics, Xi'an Institute of Optics and Precision Mechanics, Chinese Academy of Sciences, Xi'an 710119, Shaanxi, P. R. China. e-mail: xuelong\_li@opt.ac.cn.}}


\markboth{}%
{Shell \MakeLowercase{\textit{Y. Pang, et al.}}: Learning Sampling Distributions for Efficient Object Detection}

\maketitle

\begin{abstract}
 Object detection is an important task in computer vision and learning systems. Multistage particle windows (MPW), proposed by Gualdi \textit{et al}., is an algorithm of fast and accurate object detection. By sampling particle windows from a proposal distribution (PD), MPW avoids exhaustively scanning the image. Despite its success, it is unknown how to determine the number of stages and the number of particle windows in each stage. Moreover, it has to generate too many particle windows in the initialization step and it redraws unnecessary too many particle windows around object-like regions. In this paper, we attempt to solve the problems of MPW. An important fact we used is that there is a large probability for a randomly generated particle window not to contain the object because the object is a sparse event relevant to the huge number of candidate windows. Therefore, we design the proposal distribution so as to efficiently reject the huge number of non-object windows. Specifically, we propose the concepts of rejection, acceptance, and ambiguity windows and regions. This contrasts to MPW which utilizes only on region of support. The PD of MPW is acceptance-oriented whereas the PD of our method (called iPW) is rejection-oriented. Experimental results on human and face detection demonstrate the efficiency and effectiveness of the iPW algorithm. The source code is publicly accessible. 
\end{abstract}

%
%
\IEEEpeerreviewmaketitle

\section{Introduction}
\IEEEPARstart{O}{bject} detection is a key component of many computer vision systems \cite{Pang_TCYB2014}, \cite{Yu_TCYB2015}. Generally, object detection consists of two steps: feature extraction and classification \cite{Yan_TCBY2015}, \cite{Shao_TNNLS2014}, \cite{Bian_TNNLS2014}, \cite{Fan_TNNLS2014}. In this paper, we divide the task of object detection into three steps: window generation, feature extraction, and classification. Window generation outputs windows determined by shape, location and size. Features are extracted from the windows and then are classified by a classifier. Suppose $N$ windows are generated and the time spent in generating the windows is $t_w $. Let $t_f $ be the time of extracting a feature vector from a window (i.e., subimage) and $t_c $ be the time of classifying the feature vector as either positive or negative class. Then 
the computation time $t$ of an object detection algorithm can be expressed as:
\begin{equation}
\label{eq1}
t = t_w + N\times t_f + N\times t_c .
\end{equation}

Usually, $t_w $ is very small and can be neglected. Obviously, it is 
desirable if $N$ is as small as possible on the condition that the detection 
rate and false positive rate are acceptable, which is the goal of our 
algorithm. 

A dominant manner of window generation is sliding window (i.e., SW) based scanning. 
Given the pixel stride and scale factor, windows are determinately generated 
from top to bottom, left to right, and small to large. This deterministic 
manner requires a very large number of windows in order to detect objects 
with high detection rate. 

Windows can also be generated in a stochastic (random) manner. The 
stochastic manner can also be categorized into active sampling \cite{Andreopoulos_AOR_IJCV_2013}. 
Recently, Gualdi \textit{et al}. \cite{Gualdi_MPW_PAMI_2012} proposed to generate the windows (called 
particle windows) by sampling from a probability density function which is 
called proposal distribution (PD). This algorithm is called multi-stage particle window with 
abbreviation MPW. The initial PD is uniform distribution, 
meaning that each candidate window has the same chance to contain the 
object. If some windows of the $N_1 $ sampled particle windows (called 
particle windows in \cite{Gualdi_MPW_PAMI_2012}) have large classifier responses, then the PD is 
updated by enhancing the positions nearby these windows. Consequently, 
when sampling from the updated PD, more windows will be generated near the 
previous particle windows. Therefore, a smaller number $ N_i(N_i < N_1 ) $ of particle windows is needed to be drawn from the updated PD, 
which is why the MPW method can use smaller number of windows to get the 
same detection rate and false positive rate as the SW method. 

Despite the great success of MPW, there is still room to improve 
it. Due to the non-negativity of the weights and density, almost all 
particle windows are sampled from the regions neighboring to the particle 
windows obtained in the previous stages. Therefore, if the number of initial 
particle windows is not large enough to contain true positives, then there 
is a very large probability that MPW will not sample the positive windows 
any more. In addition, MPW generates unnecessary too many 
particle windows around the object and object-like regions. Considering that 
classification of these regions is more time-consuming than 
obvious non-object regions in the algorithm of cascade AdaBoost, so 
generating too many particle windows around the object and object-like regions greatly 
limits its efficiency. Importantly, it is unknown to use how many particle 
windows in each stage. 

In this paper, we propose to improve MPW with the aim of sampling a smaller 
number of particle windows without any loss in detection rate and false 
positive rate. In addition, we solve the problem of determining the number 
of particle windows in each stage. We call it iPW. Compared to MPW, iPW has the following characteristics and advantages. 
\begin{enumerate}
\item
iPW does not need to generate a large number of particle windows at the initial 
iteration (stage), which greatly reduces the detection time. Even if the 
initial particle windows do not contain any object, iPW can detect the 
objects at next stages.

\item
MPW draws unnecessary too many particle windows around the positive 
windows whereas iPW avoids generating the redundant particle windows by 
using the information of both rejected and accepted particle 
windows. 

\item
To use MPW, one has to empirically set the number of particle windows in 
each stage whereas this is not a problem because iPW generates a single 
particle window in each iteration (stage).

\item
iPW fully makes use of the information of rejected negative particle windows while MPW almost completely depends 
on the accepted positive particle windows. Rejected negative particle windows are 
used to directly suppress the PD around these windows and at the same time indirectly enhance the PD beyond the windows. In this sense, iPW is 
rejection-oriented while MPW is acceptance-oriented.

\item
In MPW, the uniform distribution is used in the initialization stage and 
plays an unimportant role in the later stages so that it can be omitted. In 
iPW, a dented uniform distribution is used to play an important role for 
sampling useful particle windows by rejecting impossible regions. 

\item
iPW utilizes dented Gaussian distribution while MPW utilizes full Gaussian 
distribution for sampling particle windows. By using dented Gaussian 
distribution, iPW avoids drawing many unnecessary particle windows around the object and object-like regions.

\item
To obtain the same detection accuracy, the total number of particle windows 
in iPW is much smaller than that in MPW.
\end{enumerate}

The remainder of the paper is organized as follows: In Section 2, related 
work is discussed. Section 3 reviews the MPW algorithm. The proposed iPW 
algorithm is described in Section 4. Experimental results are given in 
Section 5 before summarizing and concluding in Section 6.

\section{Related Work}
According to (1), the computation time of an object detection algorithm is 
determined by window generation, feature extraction, classification, and the 
number of windows. Accordingly, we can categorize existing efficient object 
detection algorithms into feature-reduced, classification-reduced, and 
window-reduced types. In addition, combination of the different types of 
algorithm should also be considered.

Cascade AdaBoost plus Haar-like features can be viewed as classical combination of feature-reduced and  classification-reduced method \cite{Viola_HLF_IJCV_2004}, \cite{Paisitkriangkrai_TNNLS2014}. In 
contrast, neural network based face detection \cite{Rowley_NN_PAMI_1998} is time-consuming in 
feature extraction and classification though it has comparable detection 
accuracy. Similarly, Histogram of Oriented Gradients (HOG) plus Support 
Vector Machine (SVM) \cite{Dalal_HOG_CVPR_2005} can be improved in efficiency by using integral image and 
cascade structure. The trilinear interpolation of 
HOG can also be approximated by decomposing gradients into different angle 
planes followed by simple smoothing \cite{Pang_RCHOG_SMC_2012}. There are many efficient 
object detection algorithms using the technique of integral image for 
extracting simple but rich features. One important type of features for 
human detection is integral channel features \cite{Benenson_100F_CVPR_2012}, \cite{Dollar_ICF_BMVC_2009}. 

Designing optimal cascade structure is also an important topic for 
increasing the speed of object detection. Recent methods include \textit{crosstalk cascade } \cite{Dollar_CC_ECCV_2012},
CoBE \cite{Brubaker_CBE_IJCV_2008}, LACBoost \cite{Shen_ENC_IJCV_2013}, \cite{Wang_ATCB_NNLS_2012}, sparse decision DAGS (\textit{directed acyclic graphs}) \cite{Benbouzid_DAGs_ICML_2012}, etc.  

In addition to the integral-image based algorithm, coarse-to-fine feature 
hierarchy \cite{Dollar_FFP_PAMI_2014}, \cite{Zhang_AOD_ICCV_2007} and template matching with binary representation \cite{Hinterstoisser_GRM_PAMI_2012} are 
also feature-reduced methods. The coarse-to-fine feature hierarchy is able 
to reject the majority of negative windows by the lower resolution features 
and process a small number of windows by higher resolution features \cite{Zhang_AOD_ICCV_2007}. By 
deep analysis of statistic of multiscale features, Doll\'{a}r \textit{et al}. \cite{Dollar_FFP_PAMI_2014} developed an efficient scheme for computing feature pyramids. Liu \textit{et al}. \cite{Liu_LRPD_PR_2014} developed a probability-based pedestrian mask which can be used as 
pre-filter to filter out many non-pedestrian regions. Template matching with 
binary representation for gradient information is a promising method for 
detecting textureless objects in real time \cite{Hinterstoisser_GRM_PAMI_2012}. Owing to its elegant 
feature representation and the architecture of modern computers, template 
matching with binary representation for gradient information can use 
thousands of arbitrarily sized and shaped templates for object 
detection in very fast speed \cite{Hinterstoisser_GRM_PAMI_2012}. By setting proper sliding stride, 
features can be reused to avoid computing the features in a window 
overlapping with its neighbors \cite{Pang_EHOG_SP_2011}. The information of spatial overlap can 
also be used for image matching and recognition\cite{Alexe_ESO_NIPS_2011}. Deep learning with 
rich features hierarchy is also a promising direction \cite{Girshick_RFH_Tech_2013}. 

Classifier-reduced method arrives at high efficiency by designing efficient 
classifier. In addition to cascade AdaBoost which uses a few of classifier 
to reject a lot of windows, one can design efficient linear and nonlinear SVM to classification. Vedaldi and Zisserman proposed to use explicit, 
instead of implicit, feature maps to approximate non-linear SVM \cite{Vedaldi_EAK_PAMI_2012}. Mak 
and Kung developed a low-power SVM classifier where the scoring 
function of polynomial SVMs can be written in a matrix-vector-multiplication 
form so that the resulting complexity becomes independent of the number of 
support vectors \cite{Mak_LPSVM_ASSP_2012}, \cite{Kung_GC_ASSAP_2013}. Linear SVM classifies a feature vector by 
computing the inner product between the sum of weighted support vectors and 
the feature vector. To reduce the computation complexity of the 
inner-product based classification, Pang \textit{et al}. proposed a sparse 
inner-product algorithm \cite{Pan_ESOD_SP_2013}. The idea is that neighboring sub-images are 
also neighboring to each other in the feature space and have similar 
classifier responses. Pang \textit{et al}. also developed a distributed strategy for computing the classifier response\cite{Pang_TCYB2014}. 

Window-reduced method is a promising direction developed in recent five 
years. This kind of methods aim at reducing the number of windows where 
feature extraction and classification have to be conducted. When an image is 
represented by a small number of keypoints and their descriptors (i.e., 
visual words), branch{\&}bound (a.k.a., efficient subwindow wearch (ESS)) is 
very efficient because it hierarchically splits the parameter spaces into 
disjoint spaces and uses quality functions to reject large parts of the 
parameter space \cite{Lampert_ESS_PAMI_2009}. Branch{\&}bound can also be used in implicit shape 
model which adopts hough-style voting \cite{Lehmann_FPRISM_IJCV_2011}, \cite{Leibe_ICS_IJCV_2008}. Branch{\&}rank generalizes 
the idea of branch{\&}bound by learning a ranking function that prioritises 
hypothesis sets that do contain an object over those that do not \cite{Lehmann_BB_IJCV_2013}. 
Acting testing is also a promising method for rapid object detection 
\cite{Andreopoulos_AOR_IJCV_2013}, \cite{Sznitman_AT_PAMI_2010}. But these visual-word based methods are not suitable for detecting 
textureless objects because it is not reliable to detect keypoints from the 
objects. 

As a signal can be reconstructed from irregularly sampled points \cite{Pang_TruncationError_TCBY2015}, an object can be detected by random sampling a fraction of all the possible locations and scales. Multi-stage particle window (MPW) is a window-reduced object detection method \cite{Gualdi_MPW_PAMI_2012} using random sampling. Branch{\&}rank 
and branch{\&}bound are based on keypoint detection whereas 
MPW extracts Haar-like features, HOG, or other features as the same manner of sliding window based object detection method. 
Therefore, MPW is expected to be suitable for detecting both texture-rich and 
texture-less objects. Sliding window based method investigates all the windows 
overlappingly and uniformly spaced in spatial and scale domains. In 
contrast, MPW only checks the windows generated from an updating proposal 
distribution. As iteration proceeds, the main peaks of the distribution 
evolve towards the objects. The open problem in MPW is how 
many windows should be generated in each iteration (stage). Existing MPW 
uses empirical numbers which is hard to result in optimal solution. In this 
paper, we propose a novel particle window based object detection method that 
is more efficient than MPW and avoids choosing particle numbers in multiple 
stages.

It is noted that the technique of Detection Proposals (DP) \cite{Zitnick_ECCV2014} (sometimes called Objectness \cite{Cheng_BING_CVPR2014} or Selective Search \cite{Uijlings_IJCV2013}) also generates a number of windows by sampling from all the possible windows. But the number of generated windows has to be large (e.g., 10\^3 or 10\^4) enough if acceptable detection quality is required. Moreover, the time spend on generating detection proposals is not satisfying (see Table 2 of \cite{Hosang_HowGood_BMVC2014}). Nevertheless, the DP methods have been successfully employed in deep leaning based detection \cite{Tian_CVPR2015}.

\section{Multi-stage Particle Windows}
The algorithm of multi-stage particle window (MPW) \cite{Gualdi_MPW_PAMI_2012} is the basis of our method. 
\subsection{Algorithm}
MPW investigates a fraction of all candidate windows in an image by sampling from a 
proposal distribution. Each particle window represents a window ${\bf 
w}=(x,y,s)^T$ where $x$, $y$ and $s$ are the horizontal position, the 
vertical position and the size of the window, respectively. The window can 
also be expressed as ${\bf w}(x,y,s)$. Once a window ${\bf w}$ 
is generated, the feature vector is then extracted and classified. Let $f(
{\bf w})$ be the classifier response. The main issue of MPW is how to design the proposal 
distribution.

At the beginning of the MPW algorithm, no prior knowledge is known about the 
preference on the candidate windows. So the proposal distribution $q_0 ( 
{\bf w}) = q_0 (x,y,s)$ is modeled as a uniform distribution $u( {\bf 
w}) = u(x,y,s) = 1 / N$, where $N$ is 
number of all possible windows in the image. 

In the first iteration, $N_1 $ particle windows are drawn from the uniform proposal 
distribution $q_0 ( {\bf w}) = u({\bf w})$ (see Fig. 1(a)). The classifier response $f( {\bf w}_i )$ is 
normalized by 
\begin{equation}
\label{eq2}
f({{\bf w}}_i ) \leftarrow \frac{f({{\bf w}}_i )}{\sum\nolimits_{j = 
1}^{N_1 } {f({{\bf w}}_j )} },
\end{equation}
so that $\sum\nolimits_{i = 1}^{N_1 } {f( {\bf w}_i )} = 1$, and $0 \le 
f({\bf w}_i ) \le 1$. Then the proposal distribution is updated 
according to the classifier responses $f( {\bf w})$ of the $N_1 $ 
particle windows:
\begin{equation}
\label{eq3}
q_1 ({{\bf w}}) {=} (1 - \alpha _1 )q_0 ( {\bf w}) {+} \alpha _1 
\sum\nolimits_{j = 1}^{N_1 } {f({\bf w}_j )} G( {\bf w}_j , 
\Sigma ).
\end{equation}

In (\ref{eq3}), $G( {\bf w}_j , \Sigma)$ is a Gaussian distribution 
where the mean is centered at $ {\bf w}_j $ and $  \Sigma$ 
is the covariance of Gaussian distribution. The weight $\alpha _1$ balances 
the previous proposal distribution $q_0$ and the mixture of Gaussian 
distributions. Gualdi \textit{et al}. \cite{Gualdi_MPW_PAMI_2012} found that $\alpha _1 = 1$ is almost the 
best choice, meaning the unimportance of previous proposal distribution. In 
Section 3.2 we will explain why $\alpha _1 = 1$ is a reasonable choice.

The sum term in (\ref{eq3}) is called measurement density function $p_1 ( {\bf 
w})$ \cite{Gualdi_MPW_PAMI_2012} :
\begin{equation}
\label{eq4}
p_1 ({\bf w}) = \sum\nolimits_{j = 1}^{N_1 } {f({\bf w}_j )G( {\bf w}_j, \Sigma )} .
\end{equation}

If $\alpha _1 = 1$, then the proposal distribution $q_1 ({\rm {\bf w}})$ is 
identical to the measurement density function $p_1 ({\rm {\bf w}})$.

In the second iteration, $N_2 $ particle windows are drawn from $q_1 ( 
{\bf w})$. Usually, $N_2$ is smaller than $N_1 $. Because the classifier 
response $f({\bf w})$ is large in the regions nearby the positives, 
most of the $N_2$ particle windows lie around the positives (see Fig. 
1(b)). 

The iteration continues with the new proposal distribution in stage $i$:
\begin{equation}
\label{eq5}
q_i ({\bf w}) {=} (1 - \alpha _i )q_{i - 1} ({\bf w}) {+} \alpha _i 
\sum\nolimits_{j = 1}^{N_i } {f( {\bf w}_j )} G({\bf w}_j, \Sigma_i).
\end{equation}
If $\alpha _i = 1$, the proposal distribution becomes:
\begin{equation}
\label{eq6}
q_i ( {\bf w})= p_i ({\bf w}) = g_i ( {\bf w}) = \sum\nolimits_{j = 1}^{N_i } 
{f({\bf w}_j )} G( {\bf w}_j ,  \Sigma _i),
\end{equation}
which is in fact the employed proposal distribution in the experiments in 
\cite{Gualdi_MPW_PAMI_2012}.

Algorithm 1 shows the procedure of MPW.

\begin{figure}[!t]
\label{Fig1}
\centering
\subfigure[]{\label{Fig1:a}
\includegraphics[width=1in]{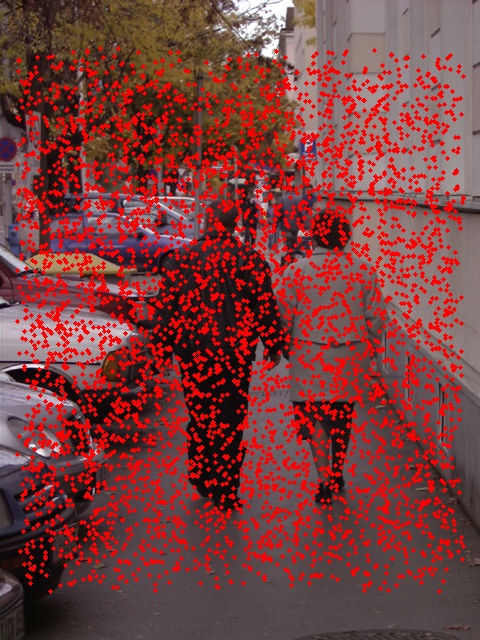}}
\hfil
\subfigure[]{\label{Fig1:b}
\includegraphics[width=1in]{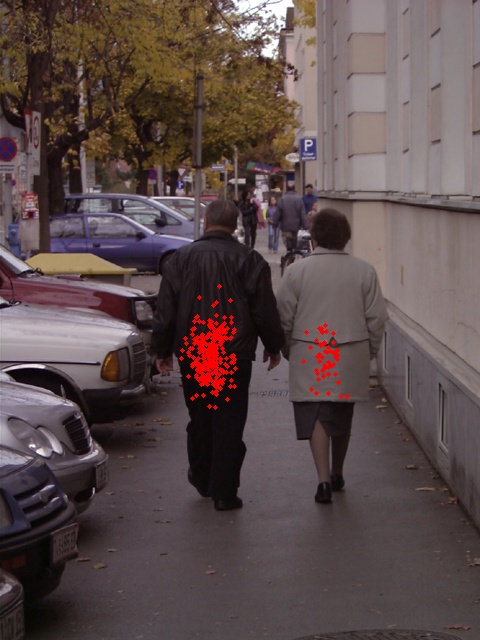}}
\caption{The process of MPW.
\subref{Fig1:a} Stage 1 samples initial particle windows by uniform distribution.
\subref{Fig1:b} Stage 2 generates particle windows around pedestrian.}
\end{figure}

\begin{algorithm}[!t]
\renewcommand{\algorithmicrequire}{\textbf{Input:}}
\renewcommand\algorithmicensure {\textbf{Output:} }
\renewcommand\algorithmicrepeat {\textbf{Iteration} }
\caption{The algorithm of MPW.}
\begin{algorithmic}[1]
\REQUIRE ~~\\
Stage number $S$;
\\The number $N_i $ of particle windows in stage $i$, $i = 1,\ldots,S$;
\\ The number $N$ of all candidate windows.
\ENSURE ~~\\ 
The set ${\bf W}_P $ of positive particle windows.
\STATE \textbf{Initialization}

\STATE Empty the set of positive particle windows: ${\bf W}_P \leftarrow \Phi $.

\STATE Initialize the proposal distribution $g({\bf w})$ by uniform distribution $u( {\bf w})\mbox{ = }1 / N$, i.e., $g( {\bf w}) \leftarrow 1 / N$.

\FOR{$s=1$ \TO $S$}
\STATE Sample $N_i $ particle windows from $g({\bf w})$.

\STATE Put the $N_i $ particle windows into ${\bf W}$, i.e., ${\bf W}\mbox{ = \{} {\bf w}_1 ,\ldots, {\bf w}_{N_i } \mbox{\}}$.

\STATE If $f( {{\bf w}_j})\mbox{ = }1$, $j\mbox{=}1,\ldots,N_i$, then ${\bf W}_P \mbox{ = } {\bf W}_P \cup  {\bf w}_j $.

\STATE Update $g( {\bf w})$ using the $ {\bf W}$, and empty ${\bf W}$.
\ENDFOR

\RETURN ${\bf W}_P$.
\end{algorithmic}
\end{algorithm}

\subsection{Why $\alpha_i = 1$}
In this section, we explain why $\alpha _i = 1$ is a reasonable choice. To the best of our knowledge, we are the first to 
explain why $\alpha _i = 1$.

For the sake of simplicity, we assume that the scale is fixed and there is only one object in the 
image. For an $h \times w$ image, the number of candidate windows is 
$M=h\times w$. Let the size of support region be $m\ll M$. In the initialization step, $N_1 $ particle windows are drawn from the 
uniform distribution $q_0 ({\bf w}) = u( {\bf w})$. Then the probability $p$ that the $N_1 $ particle 
windows contain the object is $p = 1-(1- {{m}/{M}})^{N_1}$. 
Suppose that $M=640\times 480$, $m= 50$, and $N_1 = 
1000$, then $p=0.15$. Obviously, if $N_1 $ is small, 
it is a small probability that particle windows contain the 
object. 

In the later stage, MPW generates a smaller number of particle windows, where $(1 
- \alpha _1 )$ fraction is from the uniform distribution. The probability that the $(1 - 
\alpha _1 )$ fraction of particle windows contains the object is much smaller. So 
$\alpha _i $ is usually set 1.

\subsection{Merits and Limitations}
Compared to SW, MPW is able to obtain similar accuracy at lower 
computational load \cite{Gualdi_MPW_PAMI_2012}. However, it is not clear that how to optimally select the number $m$ of 
stages and the number $N_i $ of particle windows in each stage. Guadldi \textit{et al}. 
gives an empirical rule for parameter selection:
\begin{equation}
\label{eq8}
N_i = N_1 \times e ^{-\gamma(i - 1)}, i = 1,...,m,
\end{equation}
where $N_1 $ is the initial number of particle windows. The empirical values 
of $m$ and $\gamma $ are 5 and 0.44, respectively. The exponential rule of (\ref{eq8}) makes the number $N_i $ decrease from stage to stage. Representative values of $N_i $ are given in Table 1.

\begin{table}[!t]
\centering
\renewcommand{\arraystretch}{1.2}
\caption{Representative values of $N_i $}
\begin{tabular}
{|c|c|c|c|c|c|}
\hline
Stage number $i$& 1& 2& 3& 4& 5 \\
\hline
$N_i $& 2000& 1288& 829& 534& 349 \\
\hline
\end{tabular}
\end{table}

There are three problems with the MPW algorithm:

\begin{enumerate}
\item
$N_1 $ has to be large enough so that the $N_1 $ particle windows to 
some extent overlap the objects in the image. Otherwise, $N_2 ,\ldots, N_m $ particle windows in later stages are hard to detect the objects. 
Extremely, if none of the number $N_1 $ of particle windows are positives, 
then the subsequent $N_2 $ new particle windows sampled from $q_1 ( {\bf w}) = 
\sum\nolimits_{i=1}^{N_1} {f( {\bf w}_i )G( {\bf 
w}_i, \Sigma )} $ will not contain any clue of the location 
about the objects because $G({\bf w}_i,\Sigma )$ is 
meaningless in this case.

\item
MPW generates too many unnecessary particle windows around the object 
and object-like regions. Considering that classification of these regions is more time-consuming than obviously non-object regions 
in the algorithm of cascade AdaBoost, generating too many particle windows 
around the object and object-like regions greatly limits its efficiency.

\item 
The rule of parameter selection is not guaranteed to be optimal, because there is no reason to support that the values in Table 1 are the best. 
 
\end{enumerate}

\section{Improved Particle Windows (iPW)}
In this section, we propose to improve MPW in order to detect the objects in an image with a smaller number of particle windows (PWs). 

Firstly, we define several concepts: rejection particle window, acceptance particle window, and ambiguity particle window. Secondly, the concepts of regions of rejection and acceptance are introduced. Finally, we will describe iPW algorithm based on these concepts and explain why iPW is superior to MPW.

\subsection{Rejection, Acceptance, and Ambiguity Windows}
As stated in Section 3, each particle window represents a window ${\bf 
w} = (x,y,s)^T$ where $x$, $y$ and $s$ are the horizontal position, the 
vertical position and the size of the window, respectively. The particle 
window can also be expressed as ${\bf w}(x,y,s)$. In the following, we 
use ${\bf w}(x,y)$ to represent ${\bf w}(x,y,s)$ when 
$s$ is fixed.

We employ the classifier response $f({\bf w})$ and its low and high 
thresholds (i.e., $t_l $ and $t_h )$ to define rejection, acceptance, and 
ambiguity particle windows, respectively:

\begin{enumerate}
\item 
A window $\bf{w}$ is called \textbf{Rejection PW} (\textbf{RPW}) if $f(w) 
< t_l $. Rejection PW is the particle window which can be definitely 
classified as negative class due to its low classifier response.

\item 
A window $\bf{w}$ is called \textbf{Acceptance PW}\textbf{ (APW)} 
if $f( {\bf w}) \ge t_h $. Acceptance PW is the particle window 
which can be safely classified as positive class (object) because of its 
high value of classifier response.

\item 
A window $\bf{w}$ is called \textbf{Ambiguity PW} (\textbf{ABPW}) if $t_l 
\le f({\bf w}) < t_h $. One cannot classify ambiguity PW as positive 
class because its classifier response is not large enough, meanwhile 
cannot classify it as negative class because its classifier response is not 
low enough. And we call the set of the ambiguity particle windows ${\bf 
W}_{AB} $.
\end{enumerate}

\subsection{Regions of Rejection and Acceptance}
\textbf{Region of Rejection (RoR)} If a particle window $ {\bf w}$ is 
considered as a rejection PW, it can be used to securely reject a set of 
nearby windows. The centers of the nearby windows $ {\bf w}_i$ 
and $ {\bf w}$ itself are called Region of 
Rejection (RoR) of $ {\bf w}$. We 
denote the RoR by ${\bf R}_R$:
\begin{equation}
\label{eq9}
{\bf R}_{R} ( {\bf w}) = \{x,y\vert ~ \vert \vert  {\bf w}_i -  {\bf w} \vert \vert < r_R \},
\end{equation}
where the radius $r_R $ is the maximum radius
\begin{equation}
\label{eq10}
\begin{array}{c}
 r_R = \mathop {\max }\limits_{{\bf w}_i } \vert 
\vert {\bf w}_i - {\bf w} \vert \vert, ~~ 
 s.t.~f( {\bf w}_i ) < t_l  .
\end{array}
\end{equation}
$r_R $ is a function of the classifier response $f( {\bf 
w})$. In Fig. 2(d), the smaller $f({\bf w})$ is, 
the larger $r_R$ is. 

All the windows belonging to ${\bf R}_R ({\bf w})$ 
are represented as ${\bf W}_R ({\bf w})$.
\\

\textbf{Assumption 1 (Rejection Assumption)}. \textit{If a window }${\bf w}$\textit{ is classified as a rejection window due to }$f({\bf 
w}) < t_l $\textit{ then all the windows }${\bf W}_R ({\bf w})$\textit{ can be directly rejected (i.e., labeled as negatives) without the necessity of computing the classifier response }$f({\bf 
w}_i ), {\bf w}_i \in {\bf W}_R ({\bf w})$.
\\

This assumption is illustrated in 
Fig. 3(c), the red part in the image consists of regions of rejection.

\begin{figure}[!t]
\label{Fig2}
\centering
\subfigure[]{\label{Fig2:a}
\includegraphics[width=0.8in]{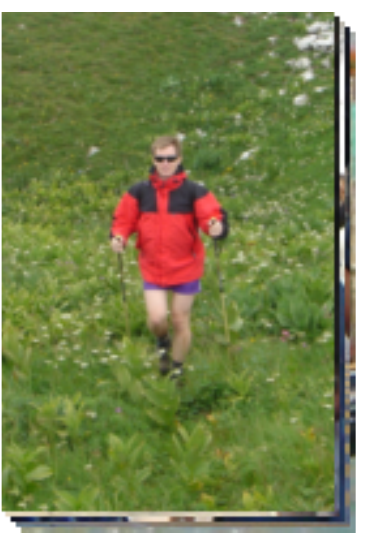}}
\hfil
\subfigure[]{\label{Fig2:b}
\includegraphics[width=1.5in]{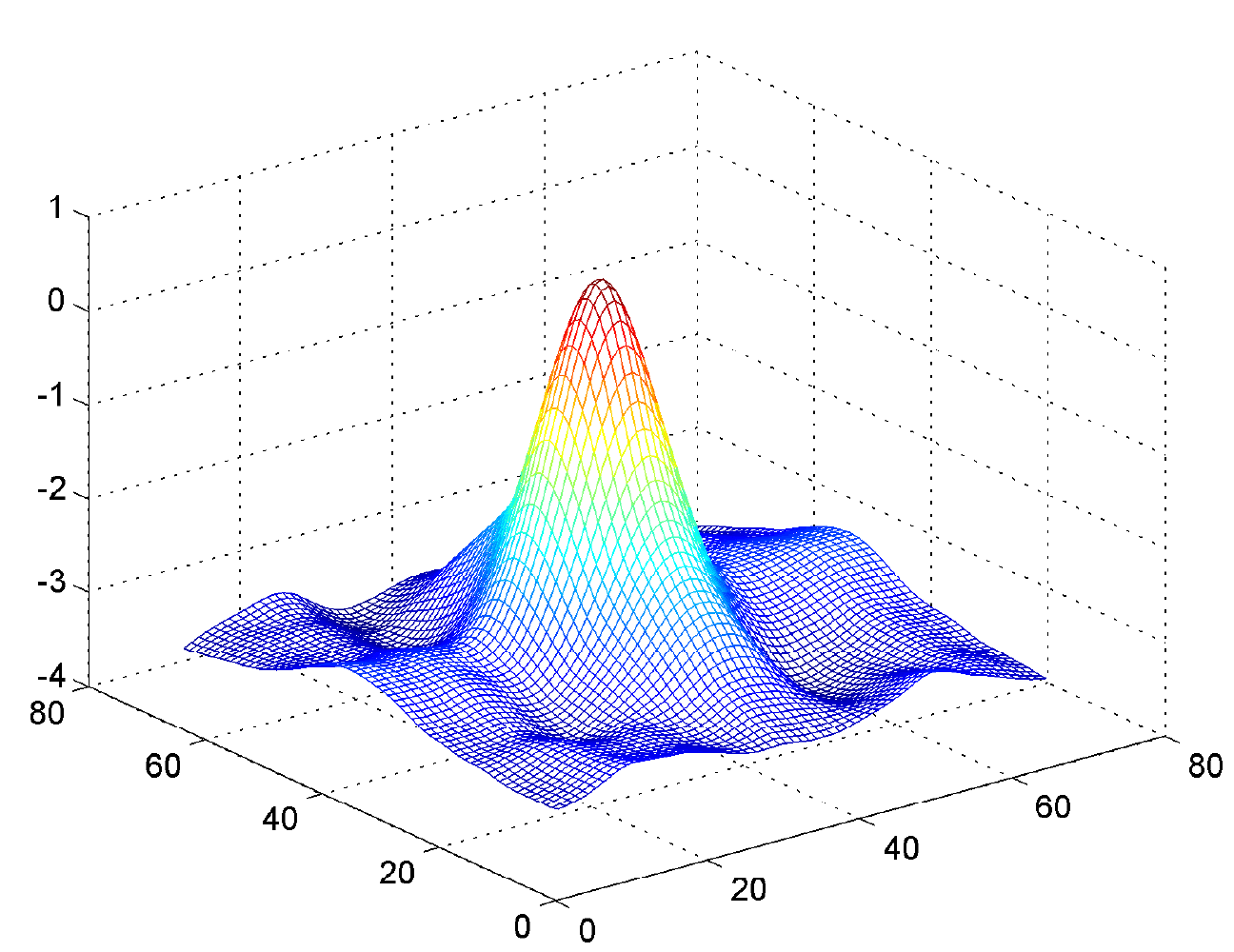}}
\hfil
\subfigure[]{\label{Fig2:c}
\includegraphics[width=1.5in]{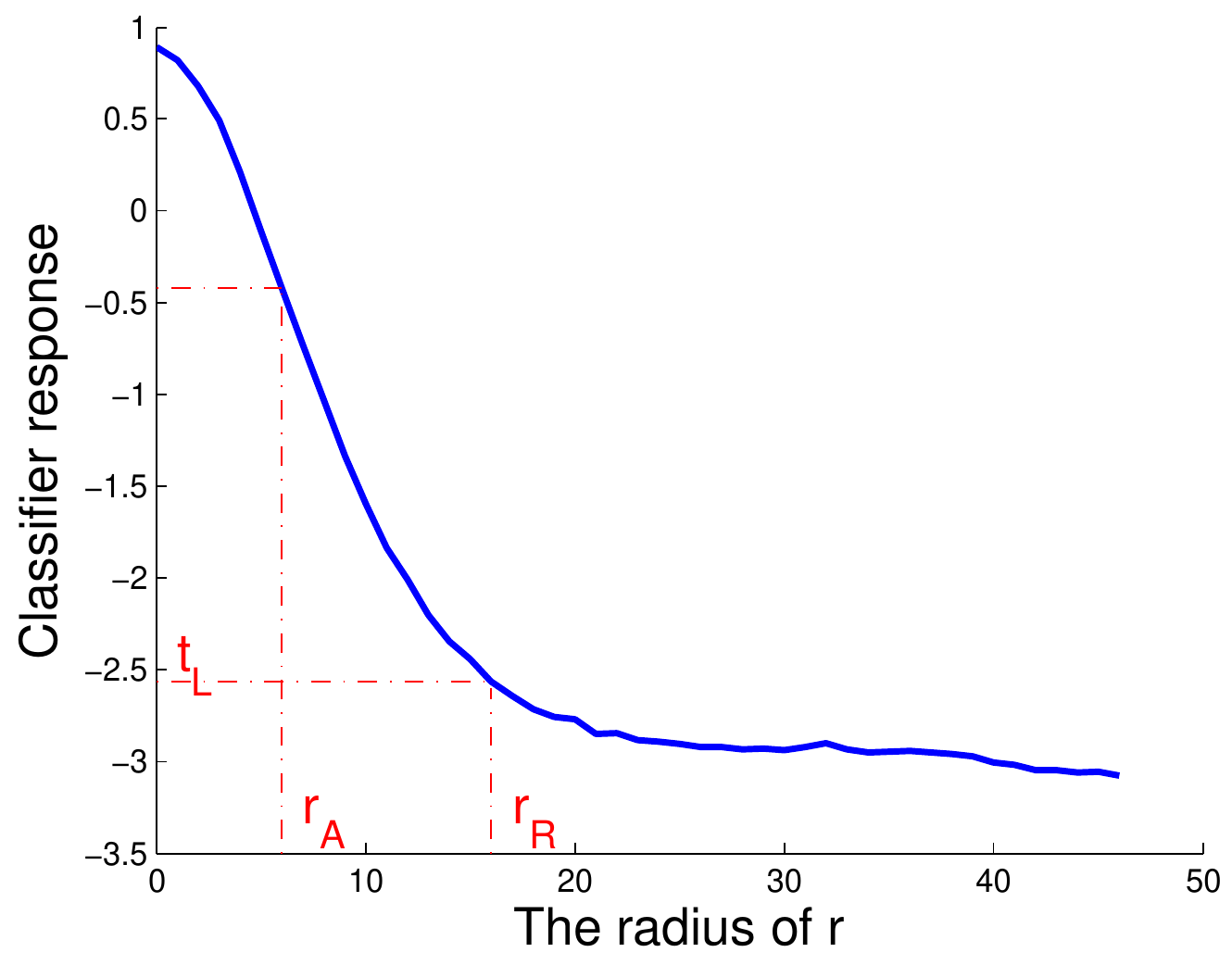}}
\hfil
\subfigure[]{\label{Fig2:d}
\includegraphics[width=1.5in]{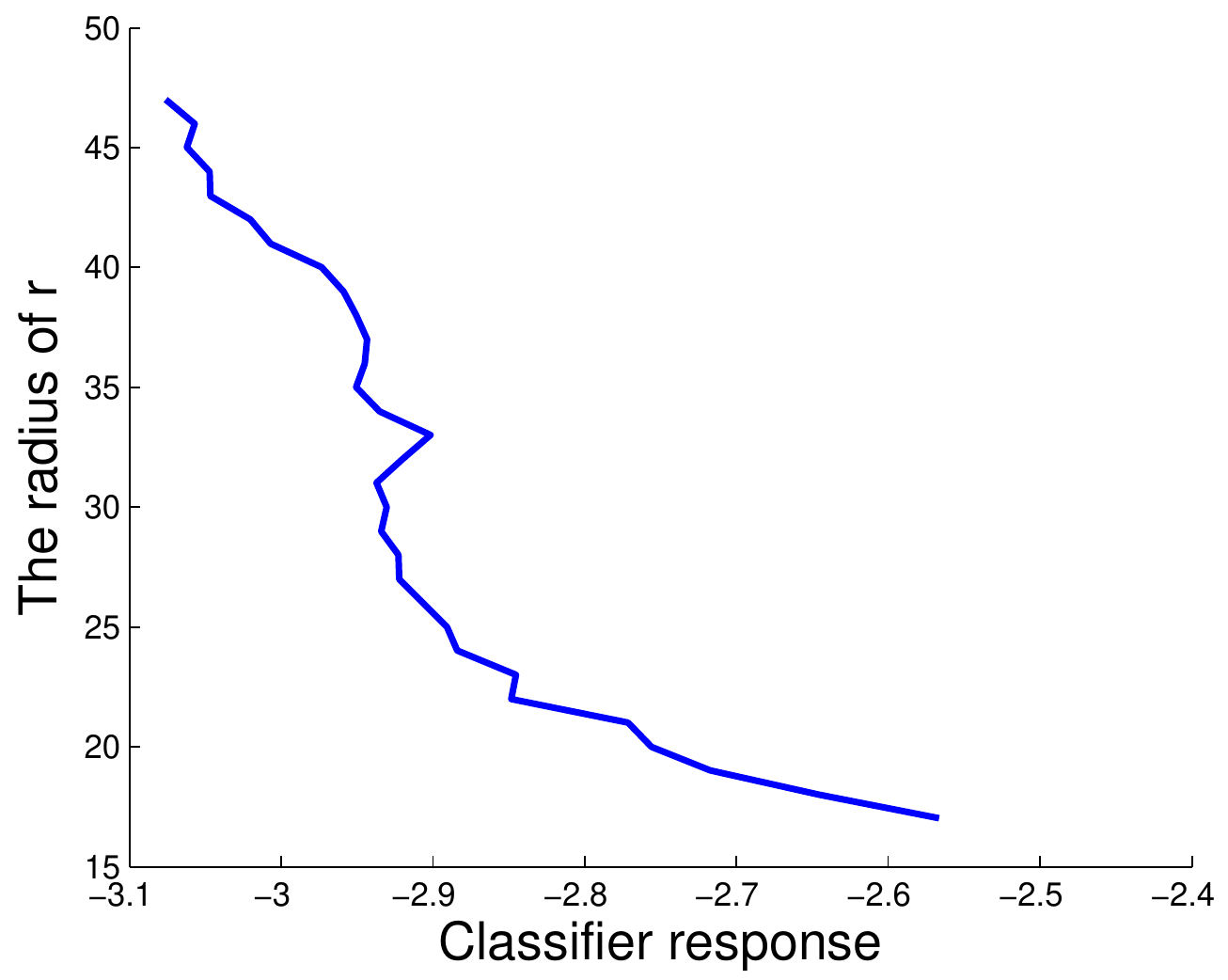}}
\caption{
\subref{Fig2:a} The original images.
\subref{Fig2:b} The average classifier response $f({\bf w}_i )$ of the windows.
\subref{Fig2:c} A profile of (b).
\subref{Fig2:d} The rejection radius $r_R $ varies with $f({\bf w}_i)$.}
\end{figure}

\textbf{Region of Acceptance (RoA)} If a particle window ${\bf w}$ is 
considered as an acceptance PW, it can be used to securely accept a set of nearby windows. The centers of the nearby 
windows ${\bf w}_i$ and ${\bf w}$ 
itself form Region of Accept (RoA) of ${\bf 
w}$. Mathematically, RoA can be denoted by ${\bf R}_A$:
\begin{equation}
\label{eq11}
{\bf R}_A ({\bf w}) = \{x,y\vert ~\vert \vert {\bf w}_i - {\bf w} \vert \vert < r_A \},
\end{equation}
where the radius $r_A $ is the maximum radius
\begin{equation}
\label{eq12}
\begin{array}{c}
 r_A = \mathop {\max }\limits_{{\bf w}_i } \vert 
\vert {\bf w}_i - {\bf w} \vert \vert, ~~ 
 s.t.~f({\bf w}_i ) \ge t_l .
 \end{array}
\end{equation}
Note that the constant is $f({\bf w}_i ) \ge t_l $ instead of $f(
{\bf w}_i ) \ge t_h $. Obviously, $r_A $ with $f({\bf w}_i ) \ge t_l 
$ is larger than $r_A $ with $f({\bf w}_i ) \ge t_h $.

All the windows belonging to ${\bf R}_A ({\bf w})$ 
are represented as ${\bf W}_A ({\bf w})$.
\\

\textbf{Assumption 2 (Acceptance Assumption)}.  \textit{If a window }$ {\bf w}$\textit{ is classified as acceptance window due to }$f( {\bf 
w}) \ge t_h $\textit{ then all the windows }${\bf W}_A ({\bf w})$\textit{ should be directly accepted (i.e., labeled as positives) without the necessity of computing the classifier response }$f( {\bf 
w}_i ), {\bf w}_i \in {\bf W}_A ( {\bf w}).$
\\

In Fig. 3(d), the blue part in the image consists of the regions of acceptance.

Let the center of a window ${\bf w}$ coincide with the center of an 
object and the size (scale) of the windows match that of the object very 
well (Fig. 2(a)). Then we compute the classifier responses of the neighboring windows 
${\bf w}_i $. As shown in Fig. 2(b), it is usual that $f( 
{\bf w})$ is the largest and $f({\bf w}_i )$ decreases with the 
distance $\vert \vert  {\bf w}_i -  {\bf w}\vert \vert $. 
In Fig. 2(c), the radius of $r_A $ is chosen so that it satisfies $f({\rm 
{\bf w}}_i ) \ge t_l $. Note that in Fig. 2(d), the rejection radius $r_R$ is derived from Fig. 2(c). 

\begin{figure}[!t]
\label{Fig3}
\centering
\subfigure[]{\label{Fig3:a}
\includegraphics[width=1.0in]{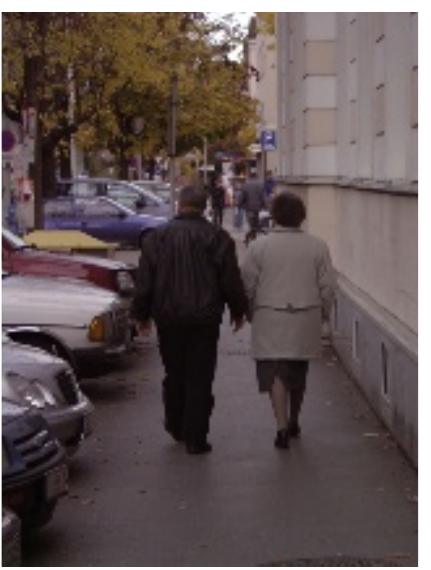}}
\hfil
\subfigure[]{\label{Fig3:b}
\includegraphics[width=1.25in]{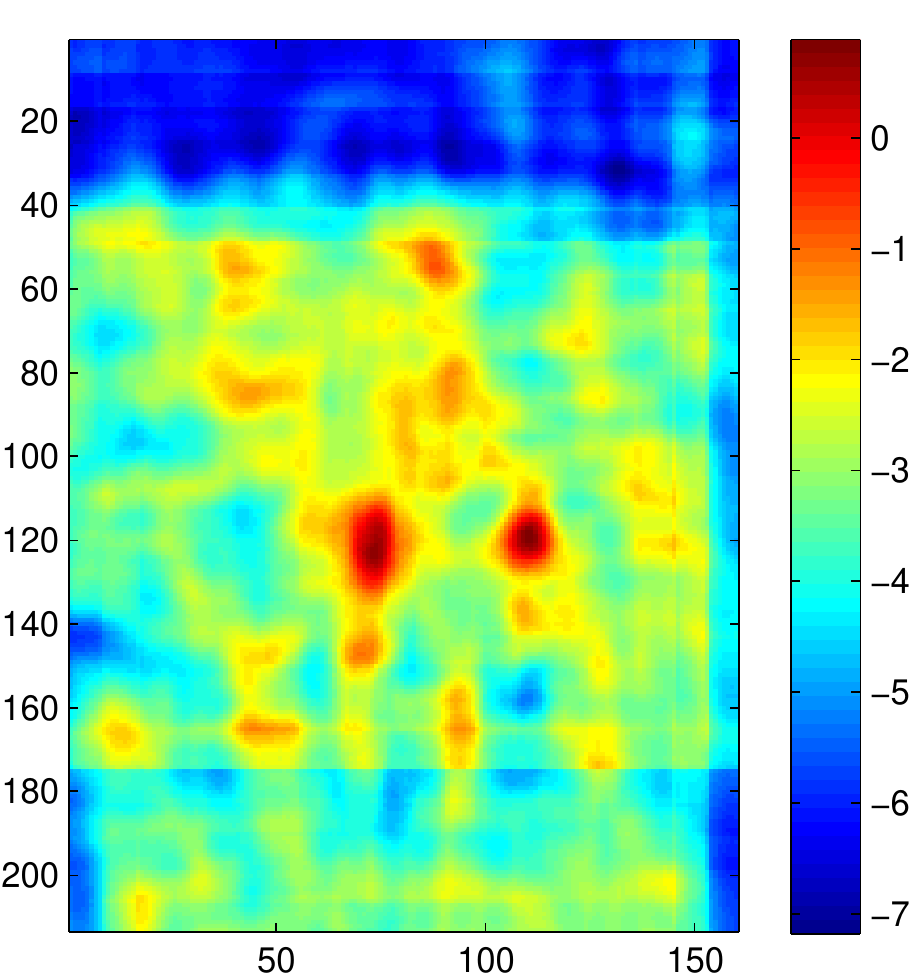}}
\hfil
\subfigure[]{\label{Fig3:c}
\includegraphics[width=1.0in]{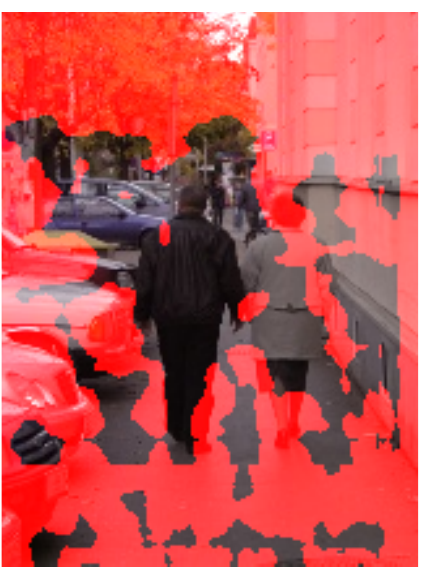}}
\hfil
\subfigure[]{\label{Fig3:d}
\includegraphics[width=1.0in]{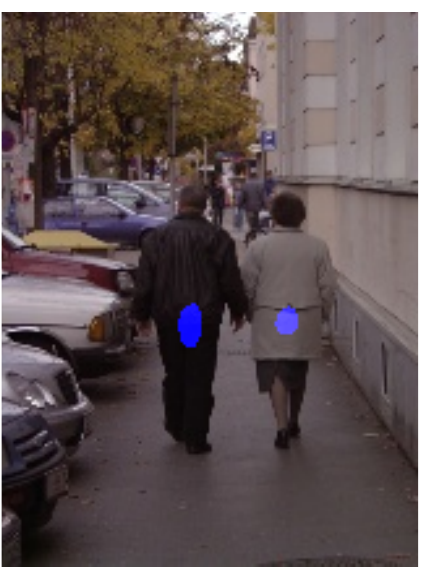}}
\caption{ Illustration of RoR and RoA.
\subref{Fig3:a} The original image.
\subref{Fig3:b} The classifier response.
\subref{Fig3:c} Regions of rejection (RoR).
\subref{Fig3:d} Regions of acceptance (RoA).}
\end{figure}

\subsection{iPW: Rejection-based Random Sampling}
In this section, we describe iPW based on the concepts of RPW, APW, and ABPW, and their corresponding RoR and RoA.

\subsubsection{Proposal Distribution of iPW}
We introduce the motivation of iPW by firstly analyzing the properties and 
limitations of MPW.

\textbf{Rejection Particle Windows and Dented Uniform Distribution }$\tilde 
{u}_R ( {\bf w})$ As discussed in Section 3.1, the particle windows of 
MPW are sampled from $\sum {f({\bf 
w}_i )G({\bf w}_i, \Sigma)} $. The contribution of a 
particle windows ${\bf w}_i $ is determined by its weight $f({\bf 
w}_i )$. If $f({\bf w}_i )$ is very small, the ${\bf w}_i $ 
contributes little to object detection because subsequent stage will not 
sample windows from the corresponding distribution $f( {\bf w}_i 
)G( {\bf w}_i , \Sigma)$. However, if only a small number 
of particle windows is generated, most of them will have small weights and the regions of objects will be not sampled . In this paper, we proposed 
how to make use of these particle windows with small weights (i.e., RPWs) for efficient object detection. One of the main contributions of the 
paper is to use rejection particle windows ${\bf w}_i $ and the 
corresponding ${\bf W}_R ( {\bf w}_i )$ (i.e., the set of windows 
insides the region ${\bf R}_R ({\bf w}_i )$) to form a dented uniform distribution $\tilde {u}_R 
( {\bf w})$. Let the number of RPWs be $N_R $, then $\tilde {u}_R ( 
{\bf w})$ is expressed as:
\begin{equation}
\label{eq13}
\tilde {u}_R ({\bf w})=
\begin{cases}
0,&{\bf w} \in \{{\bf W}_R ( {\bf w}_1 ) \cup \ldots 
\cup  {\bf W}_R ( {\bf w}_{N_R } )\}, \\
\frac{1}{a_R },&\text{otherwise},
\end{cases}
\end{equation}
where $a_R $ satisfies $\int\!\!\!\int\!\!\!\int {\frac{1}{a_R }} dxdyds = 
1$. Fig. 4(b) illustrates an one-dimensional dented uniform distribution. 
Obviously, sampling from this dented uniform distribution avoids drawing 
windows from the existing regions of rejection.

\begin{figure}[!t]
\label{Fig4}
\centering
\subfigure[]{\label{Fig4:a}
\includegraphics[width=1.5in]{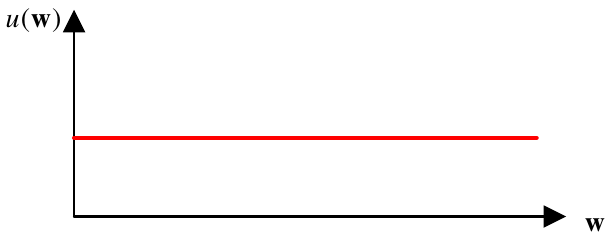}}
\hfil
\subfigure[]{\label{Fig4:b}
\includegraphics[width=1.5in]{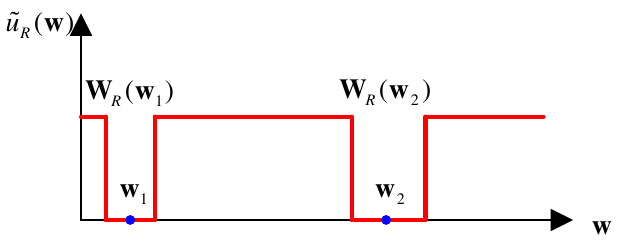}}
\hfil
\subfigure[]{\label{Fig4:c}
\includegraphics[width=1.5in]{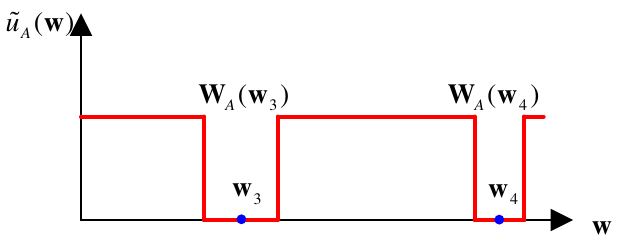}}
\hfil
\subfigure[]{\label{Fig4:d}
\includegraphics[width=1.5in]{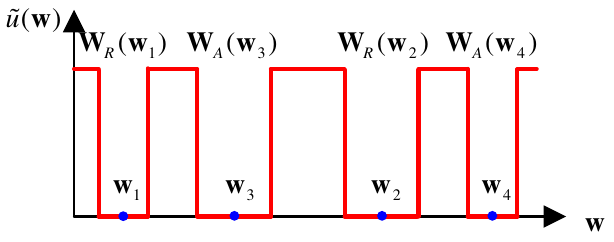}}
\caption{Uniform distribution and dented uniform distribution.
\subref{Fig4:a} Uniform distribution.
\subref{Fig4:b} Dented uniform distribution $\tilde {u}_R ( {\bf w})$ formed by two rejection particle windows $ {\bf w}_1 $ and ${\bf w}_2 $.
\subref{Fig4:c} Dented uniform distribution $\tilde {u}_A ( {\bf w})$ formed by two acceptance particle windows ${\bf w}_3 $ and ${\bf w}_4 $.
\subref{Fig4:d}  Mixture dented unform distribution by combing the rejection and acceptance particle windows.}
\end{figure}

\textbf{Acceptance Particle Window and Dented Uniform Distribution }$\tilde 
{u}_A ({\bf w})$ In MPW, when a particle window ${\bf w}_i $ 
(i.e., acceptance particle window) has large weight $f({\bf w}_i )$, 
it has large contribution to the distribution $\sum {f({\bf w}_i )G({\rm {\bf w}}_i , \Sigma)} 
$. Sampling from this updated distribution will generate particle windows overlapping or even coinciding with ${\bf w}_i $. We think that it is redundant to resample the 
acceptance window ${\bf w}_i $ and its close neighbors. To avoid the 
redundancy, we propose to employ the acceptance particle windows to maintain and update a dented uniform 
distribution $\tilde {u}_A ({\bf w})$:
\begin{equation}
\label{eq14}
\tilde {u}_A ({\bf w})=
\begin{cases}
0,&{\bf w} \in \{ {\bf W}_A ({\bf w}_1 ) \cup ... 
\cup {\bf W}_A ({\bf w}_{N_A } )\}, \\
\frac{1}{a_A },&\text{otherwise},
\end{cases}
\end{equation}
where $a_A$ satisfies $\int\!\!\!\int\!\!\!\int {\frac{1}{a_A}} dxdyds = 1$, and 
$N_A $ is the number of acceptance particle windows. Fig. 4(c) illustrates a 
$\tilde {u}_A ({\bf w})$.

Obviously, both the rejection and acceptance particle windows play role in 
excluding regions in uniform distribution. Therefore, as illustrated in Fig. 
4(d), we propose to combine $\tilde {u}_R ({\bf w})$ and $\tilde {u}_A 
({\bf w})$ into a unified dented uniform distribution $\tilde {u}( 
{\bf w})\mbox{ = }\tilde {u}_R ( {\bf w})\times \tilde {u}_A ( 
{\bf w})$: 
\begin{equation}
\label{eq15}
\tilde {u}( {\bf w})=
\begin{cases}
0,&{{\bf w} \in \{ {\bf W}_R \cup  {\bf W}_A \}}, \\
\frac{1}{a },&\text{otherwise},
\end{cases}
\end{equation}
where $a$ satisfies $\int\!\!\!\int\!\!\!\int {\frac{1}{a}} dxdyds = 1$.

\textbf{Ambiguity Particle Windows and Dented Gaussian Distribution }Suppose 
that there are $N_{AB} $ ambiguity particle windows in ${\bf{W}}_{AB} $. With the 
class label being either positive or negative, the region nearby the 
ambiguity particle window exhibits the largest uncertainty relative to the 
rejection particle windows and the acceptance particle windows, so it 
contains potential clue for objects. One way to exploit the $N_{AB} $ 
ambiguity particle windows is directly using them for modeling a mixture of 
Gaussian distribution $g({\bf w})$: 
\begin{equation}
\label{eq16}
g({\bf w})= \sum\nolimits_{i = 1}^{N_{AB} } {f( {\bf w}_i )} 
G({\bf w}_i ,\Sigma).
\end{equation}

However, our experimental results show that sampling from $g({\bf w})$ 
may result in particle windows overlapping with the existing rejection or 
acceptance particle windows. Clearly, it is useless to sample such particle 
windows. Therefore, to remove the redundancy, we propose to model a mixture 
of dented Gaussian distribution $\tilde {g} ({\bf w})$ using the $N_{AB}$ ambiguity 
particle windows with the help of the existing rejection and acceptance 
particle windows or equivalently the current dented uniform distribution $\tilde {u} ({\bf w})$:
\begin{equation}
\label{eq17}
\tilde {g}({\bf w})= \sum\nolimits_{i = 1}^{N_{AB} } {f( {\bf w}_i 
)\left[ {G( {\bf w}_i , \Sigma )\times \left( {a\times \tilde {u}( {\bf w})} \right)} \right]}.
\end{equation}

Because $a\times \tilde {u}( {\bf w}) = 0$ in the regions of rejection and acceptance and $a\times \tilde {u} ( {\bf w}) = 1$ elsewhere, 
so multiplying $G({\bf w}_i , \Sigma)$ with $a\times 
\tilde {u}({\bf w})$ results in dented Gaussian distribution as is 
illustrated in Fig. 5. In Fig. 5(a) two ambiguity particle windows $ 
{\bf w}_1 $ and ${\bf w}_2 $ in ${\bf W}_{AB} $ form a mixture 
of Gaussian distribution $g( {\bf w})$. In Fig. 5(b) a rejection 
particle window ${\bf w}_R $ and an acceptance particle window $
{\bf w}_A $ form a dented uniform distribution $\widetilde{u}({\bf 
w})$. Fig. 5(c) shows the final dented Gaussian ditribution $\tilde 
{g}({\bf w})$.

\begin{figure}[!t]
\label{Fig5}
\centering
\subfigure[]{\label{Fig5:a}
\includegraphics[width=1.5in]{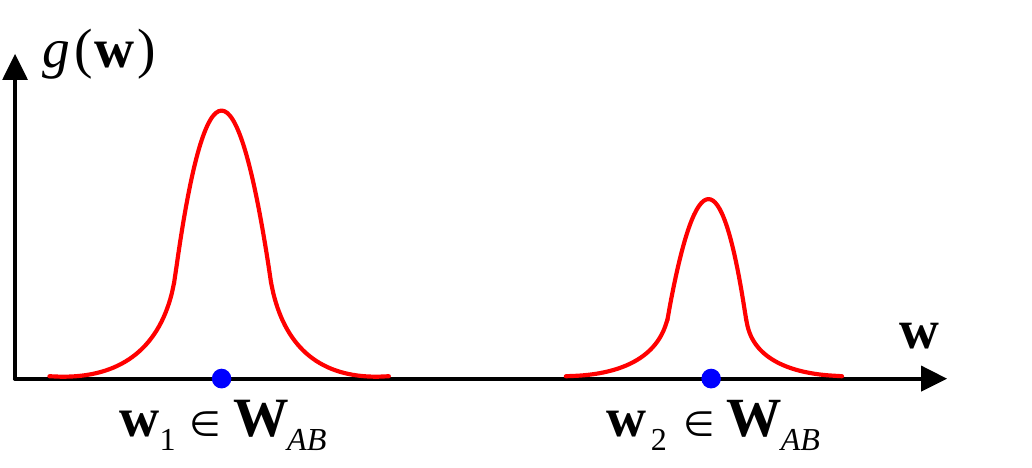}}
\hfil
\subfigure[]{\label{Fig5:b}
\includegraphics[width=1.5in]{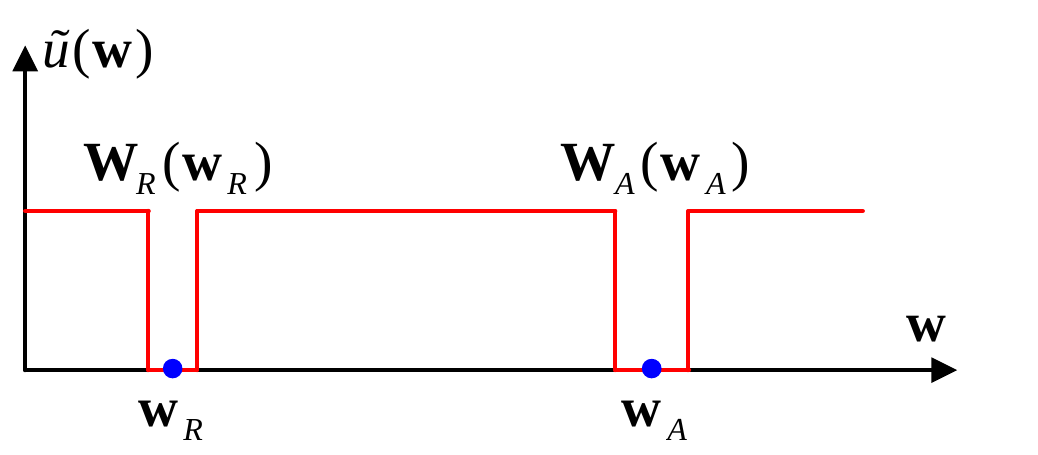}}
\vfil
\subfigure[]{\label{Fig5:c}
\includegraphics[width=1.5in]{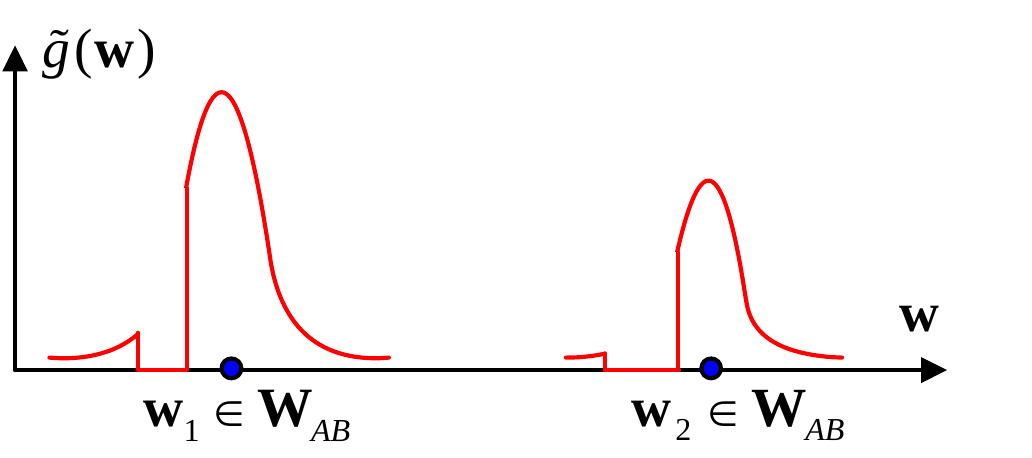}}
\caption{Dented Gaussian distribution.
\subref{Fig5:a} Gaussian distribution.
\subref{Fig5:b} Dented uniform distribution.
\subref{Fig5:c} Mixture of dented Gaussian distribution.}
\end{figure}

\textbf{Proposal Distribution}. The proposal distribution $\tilde {q}({\bf w})$ of iPW is a weighted average of the dented uniform distribution $\tilde {u}( {\bf w})$ and mixture of dented Gaussian distribution $\tilde {g}({\bf w})$: 
\begin{equation}
\begin{split}
\label{eq19}
\tilde {q}({\bf w})&= P_u \times \tilde {u}({\bf w}) + P_g 
\times \tilde {g}({\bf w})
\\&\mbox{ = }P_u \times \tilde {u}({\bf w})\\
& + P_g \sum\limits_{i = 
1}^{N_{AB} } {f({\bf w}_i )\left[ {G( {\bf w}_i , \Sigma 
)\times \left( {a\times \tilde {u}({\bf w})} \right)} \right]}.
\end{split}
\end{equation}

More generally, the proposal distribution $\tilde {q}_i ( {\bf w})$ in stage $i$ is expressed as 
\begin{equation}
\begin{split}
\label{eq20}
\tilde {q}_i ({\bf w})&\mbox{ = }P_u (i)\times \tilde {u}_i ({\bf w}) 
\\&+ P_g 
(i)\sum\limits_{j = 1}^{N_{AB} } {f({\bf w}_j )\left[ {G({\bf 
w}_j , \Sigma)\times \left( {a\times \tilde {u}_i ( {\bf 
w})} \right)} \right]}.
\end{split}
\end{equation}

The weights $P_u $ and $P_g $ can be regarded as the posterior probabilities for $ {\bf w}_i$ to be generated from $\tilde {u}({\bf w})$ and $\tilde {g}({\bf w})$, respectively. That is $p(\tilde {u}({\bf 
w})\vert  {\bf w}_i )\mbox{ = }P_u $ and $p(\tilde {g}({\bf 
w})\vert  {\bf w}_i )\mbox{ = }P_g $ which can be respectively defined by 
\begin{equation}
\label{eq21}
P_u \mbox{ = }\alpha \times \mbox{(1} - \frac{N_R + N_A }{N}),~\text{and}~P_g \mbox{ = } 1 - P_u.
\end{equation}
In (\ref{eq21}), $\alpha \in [0,1] $ is used for performance adjusting, $N_R = \vert  {\bf W}_R \vert $ and $N_A = \vert  {\bf W}_A \vert $ are 
the numbers of rejection and acceptance particle windows in $ {\bf W}_R$ and ${\bf W}_A $, respectively.

\textbf{Hypothesis 1}. Suppose there are two stages and the number of particle windows in stage 1 and stage 2 are $N_1 $ and $N_2 $, respectively. In both MPW and iPW, the $N_1 $ particle windows are generated from the same uniform distribution. But the $N_2 $ particle windows in stage 2 of MPW are 
sampled from $q( {\bf w})$ whereas they are sampled from $\tilde 
{q}( {\bf w})$ in iPW. Then the probability for $N_2 $ particle windows in iPW to contain the object is larger than that in MPW.

It is trivial to prove Hypothesis 1. If there are non-zero number of rejection 
and/or acceptance particle windows, then the search region is reduced by 
these particle windows (equivalently, the dented uniform distribution 
$\tilde {u}_1 ( {\bf w}))$. Consequently, sampling the same number of 
particle windows from reduced search domain is better than from the original 
large domain in the sense of detecting the objects. Generally, if both iPW 
and MPW use the same number of particle windows in each stage, then the 
probability for iPW to detect the objects is larger than that of MPW.

\textbf{Each Stage Consists of One Particle Window } So far, we have 
designed the new proposal distribution of iPW. The question is that how many 
particle windows are to be generated in each stage. In MPW, the exponential 
rule of (7) is used for setting the window number. But this is far from 
optimal. Intuitively, we think that it is optimal if each stage 
contains one particle window. However, it fails completely for MPW. 
Throughout the paper, iPW means the one where each stage has a single new 
particle window. The number of generated particle windows incrementally 
increases one by one. The first letter 'i' of "iPW" is named after 
"incremental".

\subsubsection{Basic Algorithm of iPW}
The core of iPW is iteratively sampling particle windows from the proposal 
distribution $\tilde {q}({\bf w})$ and updating the sets of rejection, 
acceptance, and ambiguity particle windows. Because the proposal distribution 
$\tilde {q}( {\bf w})$ is a weighted average of the dented 
distributions $\tilde {u}( {\bf w})$ and $\tilde {g}( {\bf w})$, 
drawing a particle window from $\tilde {q}({\bf w})$ is equivalent to 
drawing from either $\tilde {u}( {\bf w})$ or $\tilde {g}( {\bf 
w})$ with the probabilities $P_u $ and $P_g $, respectively. 

The basic algorithm of iPW is given in Algorithm 2. The output is ${\bf W}_P $ (the final set of positive particle windows) on which 
non-maximum-suppression is applied for final object detection. 

In the initialization step, the sets of rejection, acceptance, and ambiguity particle 
windows are emptied (line 2). The dented uniform distribution $\tilde {u}( {\bf w})$ is initialized by the uniform distribution $u({\bf w})\mbox{ = }1 / N$ because currently there are no rejection and acceptance particle windows (line 3). 
The mixture of dented Gaussian distribution $\tilde {g}( {\bf w})$ is 
initialized to be 0 because so far there are no ambiguity particle windows to really construct it (line 3).  

In each iteration, a particle window is generated from either $\tilde 
{u}( {\bf w})$ or $\tilde {g}({\bf w})$ until the predefined 
number $N_{iPW} $ of particle windows is obtained. According to the value 
$f( {\bf w})$ of classifier response, the generated particle window 
${\bf w}$ is classified as rejection, acceptance, or ambiguity 
particle window. If $ {\bf w}$ is classified as rejection particle 
window, then all the windows ${\bf W}_R ( {\bf w})$ in the $
{\bf R}_R ({\bf w})$ (region of rejection particle window of $ 
{\bf w})$ will be remerged into $ {\bf W}_R $. If ${\bf w}$ is 
classified as acceptance particle window, then all the windows ${\bf 
W}_A ({\bf w})$ in the $ {\bf R}_A ({\bf w})$ (region of 
acceptance particle window of ${\bf w})$ will be remerged into $ 
{\bf W}_A $. Otherwise it will be put into the window set ${\bf W}_{AB} 
$ (see line 9, 10 and 11).

Finally, based on the updated $ {\bf W}_R $, $ {\bf W}_A $, and 
${\bf W}_{AB} $, the dented distributions $\tilde {u}( {\bf w})$ 
and $\tilde {g}( {\bf w})$ are updated according to (14) and (16), respectively.

\begin{algorithm}[!t]
\renewcommand{\algorithmicrequire}{\textbf{Input:}}
\renewcommand\algorithmicensure {\textbf{Output:} }
\renewcommand\algorithmicrepeat {\textbf{Iteration} }
\caption{The basic algorithm of iPW.}
\begin{algorithmic}[1]
\REQUIRE ~~\\
The number $N$ of all candidate windows;
\\The number $N_{iPW}$ of total particle windows;
\\High and low classifier thresholds $t_{h }$ and $t_{l}$, respectively;
\ENSURE ~~\\ 
The set $ {\bf W}_P $ of positive particle windows.
\STATE \textbf{Initialization}

\STATE  Empty the sets of rejection, acceptance, and ambiguity particle windows: ${\bf W}_R \leftarrow \Phi $, ${\bf W}_A \leftarrow \Phi $, and ${\bf W}_{AB} \leftarrow \Phi $, respectively. Let the numbers of rejection, acceptance, and ambiguity particle windows be $N_R = 0 $, $N_A = 0$, and $N_{AB} = 0 $.

\STATE Initialize the dented uniform and dented Gaussian distributions by $\tilde {u}({\bf w}) \leftarrow 1 / N$ and $\tilde {g}( {\bf w}) \leftarrow 0$.

\STATE \textbf{Iteration:}

\FOR{$i=1$ \TO $N_{iPW}$}
\STATE $P_u = \alpha \times (1 - \frac{N_A + N_R }{N})$, $P_g = 1 - P_u $.

\STATE Sample a particle window ${\bf w}$ from either $\tilde {u}({\bf w})$ or $\tilde {g}( {\bf w})$. The probabilities for $ {\bf w}$ to be generated from $\tilde {u}({\bf w})$ and $\tilde {g}( {\bf w})$ are $P_u $ and $P_g $, respectively.

\STATE Put ${\bf w}$ into ${\bf W}_R $, ${\bf W}_A $, or ${\bf W}_{AB} $ according to the classifier response:

\STATE If $f( {\bf w}) < t_l $, then $ {\bf W}_R \mbox{ = }{\bf W}_R \cup {\bf W}_R ({\bf w})$, $N_R \leftarrow \vert {\bf W}_R \vert$;

\STATE If $f( {\bf w}) \ge t_h $, then $ {\bf W}_A \mbox{ = } {\bf W}_A \cup  {\bf W}_A ( {\bf w})$, $N_A \leftarrow \vert {\bf W}_A \vert$, and $ {\bf W}_P =  {\bf W}_P \cup  {\bf w}$;

\STATE If $t_l \le f( {\bf w}) < t_h $, then ${\bf W}_{AB} \mbox{ = }{\bf W}_{AB} \cup {\bf w}$, $N_{AB} \leftarrow \vert {\bf W}_{AB} \vert$.

\STATE Update $\tilde {u}( {\bf w})$ and $\tilde {g}( {\bf w})$ using the updated $ {\bf W}_R $, $ {\bf W}_A $, and $ {\bf W}_{AB} $.
\ENDFOR

\RETURN $ {\bf W}_P $.
\end{algorithmic}
\end{algorithm}

\textbf{Characteristics of the iPW algorithm} Fig. 6 demonstrates the 
characteristics of the iPW algorithm. Because each stage (iteration) 
generates a single particle window, the cumulative number $N_{iPW} (i)$ of 
generated particle windows at stage $i$ is $N_{iPW} (i) = i$. Among the $N_{iPW} (i)$ particle windows, $P_u (i)$ fraction is 
sampled from $\tilde {u}( {\bf w})$ whereas 
$P_g (i)$ fraction is sampled from $\tilde 
{g}({\bf w})$. That is, the number of windows coming from $\tilde 
{u}({\bf w})$ is $N_{iPW}^u (i) = P_u (i)\times N_{iPW} (i)$, and the 
number of windows coming from $\tilde {g}( {\bf w})$ is $N_{iPW}^g (i) 
\approx P_g (i)\times N_{iPW} (i)$. Fig. 6(a) shows that most of the generated particle windows are from $\tilde 
{u}({\bf w})$ in the first 
several stages. But its fraction (i.e., $P_u (i))$ decreases as 
iteration proceeds meanwhile the fraction $P_g (i)$ increases (see 
Fig. 6(b)). 

The above phenomenon is explained as follows. Because the number of object 
windows is very small relative to the total number of 
windows in the image, drawing a particle window from the initial 
distribution $u( {\bf w})$ and 
the distribution $\tilde {u}_i ({\bf w})$ in first few stages (e.g., 
$i=10$) will result in rejection particle windows in a very large probability. 
Consequently, the number $N_R (i)$ of rejection particle windows increases 
very fast with $i$, but the number $N_A (i)$ of acceptance particle windows 
increases very slowly (see Fig. 6(c)), which makes the number $N_u 
(i) = N {-} N_R (i) {-} N_A (i)$ of unvisited windows be very large for small $i$. 
According to (19), $P_u (i) \gg P_g (i)$. But as iteration 
proceeds, $N_R (i)$ and $N_A (i)$ increase monotonically making $P_u $ decrease. 

As $N_R (i)$ and $N_A (i)$ increases monotonically with 
$i$, so it can pay more attention to the other unvisited potential areas. This 
characteristic makes iPW not to generate unnecessary too many particle 
windows around the object and object-like regions. It is known that 
classification of these regions is very time-consuming if 
cascade AdaBoost is adopted. This is one of the advantages of iPW over MPW.

\begin{figure}[!t]
\label{Fig6}
\centering
\subfigure[]{\label{Fig6:a}
\includegraphics[width=1.5in]{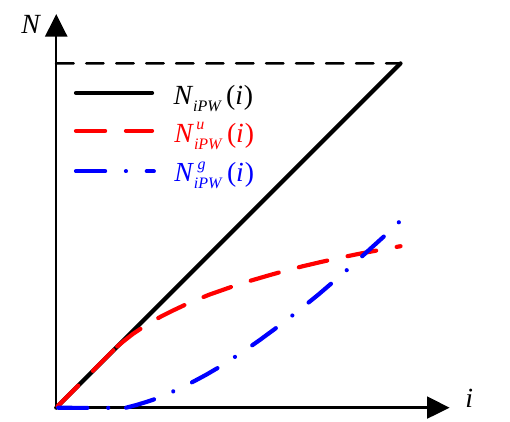}}
\hfil
\subfigure[]{\label{Fig6:b}
\includegraphics[width=1.5in]{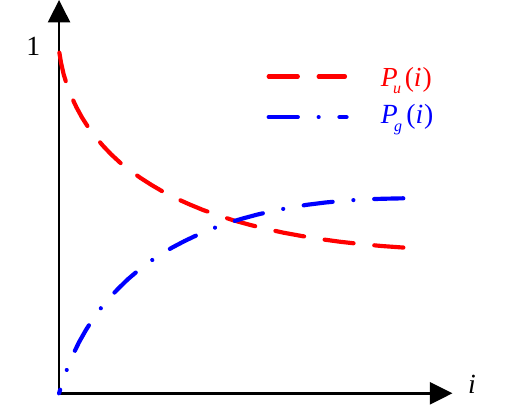}}
\vfil
\subfigure[]{\label{Fig6:c}
\includegraphics[width=1.5in]{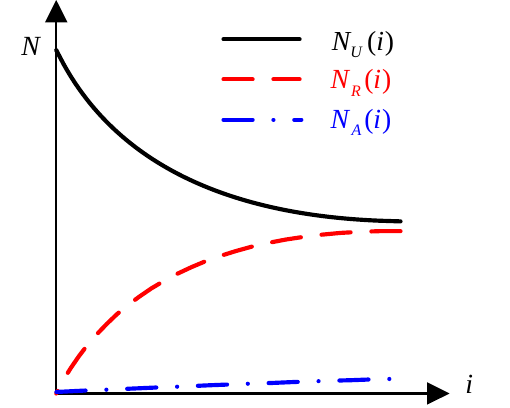}}
\caption{Characteristics of iPW.
\subref{Fig6:a} The curves of $N_{iPW} (i)$ , $N_{iPW}^g (i)$, and $N_{iPW}^u (i)$.
\subref{Fig6:b} The curves of $P_u (i)$ and $P_g (i)$.
\subref{Fig6:c} The curves of $N_U (i)$, $N_R (i)$, and $N_A (i)$.}
\end{figure}

\subsubsection{Semi-incremental Version of iPW (siPW)}
Algorithm 2 is a purely incremental algorithm which has some problems. 
First, in the first few iterations the number of particle windows is very 
small, so $N_{AB} (i)$ in it is much smaller. In this case, $\tilde {g}( 
{\bf w})$ can't reflect the probability distribution of whole image. In 
order to have enough ambiguity particles to represent the probability 
distribution, the variables $N_C^\ast $ and 
$b$ are introduced in Algorithm 3. Through them, the particle windows are forcibly sampled from $\tilde {u}( {\bf 
w})$ in first several stages until that a certain 
number $N_C^\ast $ of particle windows, especially the ambiguity particle windows, is 
available. By this way, it can better reflect the probability of whole image. Second, if a particle window is sampled from $\tilde {g}( {\bf 
w})$ and then $\tilde {g}( {\bf w})$ is immediately updated, the latter 
sampled particle windows will be heavily centered on the strongest 
classifier response regions. Namely, the regions with strongest responses will be enhanced more 
and more, while the regions with the relative 
lower responses where object exists will be ignored. So instead of updating $\tilde {g}({\bf w})$ per particle window, it is wise to update $\tilde {g}( {\bf w})$ until there is a certain number of particle windows (line 14). To overcome the above problems, a 
semi-incremental version of iPW (i.e., Algorithm 3) is proposed. The main differences 
from Algorithm 2 are written in italic. We call it semi-incremental algorithm 
because $\tilde {g}({\bf w})$ is updated 
once there is a certain number $N_C^\ast $ of pariticle windows, though each stage has one particle window and the rejection regions are updated 
in purely incremental manner.

\begin{algorithm}[!t]
\renewcommand{\algorithmicrequire}{\textbf{Input:}}
\renewcommand\algorithmicensure {\textbf{Output:} }
\renewcommand\algorithmicrepeat {\textbf{Iteration} }
\caption{Semi-incremental version of iPW.}
\begin{algorithmic}[1]
\REQUIRE ~~\\
The number $N$ of all candidate windows;
\\The number $N_{iPW}$ of total particle windows;
\\High and low classifier thresholds $t_{h }$ and $t_{l}$, respectively;
\\ {\it The threshold number $N_{C}^\ast$ for the number of ambiguity particle windows}
\ENSURE ~~\\ 
The set ${\bf W}_P $ of positive particle windows.
\STATE \textbf{Initialization}

\STATE Empty the sets of rejection, acceptance, and ambiguity particle windows: ${\bf W}_R \leftarrow \Phi $, ${\bf W}_A \leftarrow \Phi $, and ${\bf W}_{AB} \leftarrow \Phi $, respectively. Let the numbers of rejection, acceptance, and ambiguity particle windows be $N_R = 0$, $N_A = 0$, and $N_{AB} = 0$.

\STATE Initialize the dented uniform and dented Gaussian distributions by $\tilde {u}({\bf w}) \leftarrow 1 / N$ and $\tilde {g}({\bf w}) \leftarrow 0$.

\STATE {\it Initialize the binary indicator $b = 0$ and the number of cumulative particle windows $N_C = 0$}.

\STATE \textbf{Iteration:}

\FOR{$i=1$ \TO $N_{iPW}$}
\STATE  If $b = 0$, then $P_u \leftarrow 1$ and $P_g \leftarrow 0$, else $P_u = a\times (1 - \frac{N_A + N_R }{N})$ and $P_g = 1 - P_u $.

\STATE Sample a particle window ${\bf w}$ from either $\tilde {u}( {\bf w})$ or $\tilde {g}({\bf w})$ with $P_u $ and $P_g $, respectively. $N_C = N_C + 1$.

\STATE Put ${\bf w}$ into ${\bf W}_R $, ${\bf W}_A $, or ${\bf W}_{AB} $ according to the classifier response:

\STATE If $f({\bf w}) < t_l $, then ${\bf W}_R  = {\bf W}_R \cup  {\bf W}_R ( {\bf w})$, $N_R \leftarrow \vert {\bf W}_R \vert$;

\STATE If $f({\bf w}) \ge t_h $, then ${\bf W}_A \mbox{ = }{\bf W}_A \cup  {\bf W}_A ({\bf w})$, $N_A \leftarrow \vert {\bf W}_A \vert$, and $ {\bf W}_P =  {\bf W}_P \cup  {\bf w}$;

\STATE If $t_l \le f( {\bf w}) < t_h $, then $ {\bf W}_{AB} \mbox{ = }{\bf W}_{AB} \cup {\bf w}$, $N_{AB} \leftarrow \vert  {\bf W}_{AB} \vert$.

\STATE Update $\tilde {u}( {\bf w})$ using the updated ${\bf W}_R $ and ${\bf W}_A $.

\STATE {\it If $N_C = N_C^\ast $, then $b = 1$, update $\tilde {g}( {\bf w})$, ${\bf W}_{AB} \leftarrow \Phi $, $N_C^\ast = N_C^\ast \times e^{ - \gamma }$, and $N_C = 0$}.

\ENDFOR

\RETURN $ {\bf W}_P $.
\end{algorithmic}
\end{algorithm}

\subsubsection{Efficiently Sampling From Dented Uniform and Gaussian Distributions}
As can be seen from (1), the computation time of an object detection 
algorithm is composed of the time of window generation, feature extraction, 
and classification. So it is important for iPW to efficiently generating 
particle windows from $\tilde {u}({\bf w})$ and $\tilde {g}( {\bf w})$.

To efficiently draw a particle window from $\tilde {u}({\bf w})$, we, 
in Algorithm 4, propose to iteratively draw a window from standard uniform 
distribution $u( {\bf w})$ until it does not belong to $ {\bf W}_R $ 
or ${\bf W}_A $. Similarly, to efficiently draw a particle window from 
$\tilde {g}({\bf w})$, in Algorithm 4 we propose to iteratively draw a 
window from standard mixture of Gaussian distribution $g({\bf w})$ 
until it does not coincide with the elements of ${\bf W}_R $ and $ {\bf W}_A $. As one can design algorithm for checking $ {\bf w} 
\in {\bf W}_R $ and ${\bf w} \in {\bf W}_A $ in an 
extremely efficient manner, the computation time of window generation in iPW is negligible. The maximum iterations number $N_{\max } $ is used for avoiding infinite loops.

\begin{algorithm}[!t]
\renewcommand{\algorithmicrequire}{\textbf{Input:}}
\renewcommand\algorithmicensure {\textbf{Output:} }
\renewcommand\algorithmicrepeat {\textbf{Iteration} }
\caption{Draw a particle window ${\rm {\bf w}}$ from $\tilde {u}({\rm {\bf w}})$ (or $\tilde {g}( {\bf w})$).}
\begin{algorithmic}[1]
\REQUIRE ~~\\
The sets of rejection and acceptance particle windows: ${\bf W}_R $ and ${\bf W}_A $, respectively;
\\The maximum iteration number $N_{\max }$;
\ENSURE ~~\\ 
a particle window ${\bf w}$.

\FOR{$n=1$ \TO $N_{\max}$}
\STATE Draw a window ${\bf w}_n$ from the uniform distribution $u({\rm {\bf w}})$ (or $g( {\bf w})$).

\STATE  If ${\bf w} \notin {\bf W}_R $ and ${\bf w} \notin {\bf W}_A $, then ${\bf w} = {\bf w}_n$, and break.

\ENDFOR

\RETURN ${\bf w}$.
\end{algorithmic}
\end{algorithm}

%
%

\section{Experimental Results}
\subsection{Experimental Setup}
Experiments are carried out on the INRIA pedestrian dataset and the MIT-CMU face dataset to compare the proposed iPW with MPW and SW. To detect pedestrians in INRIA dataset, HOG and SVM \cite{Dalal_HOG_CVPR_2005} are 
used for features and classifier, respectively. Haar-like features and 
cascade AdaBoost classifier are employed for detecting faces in the MIT-CMU dataset \cite{Viola_HLF_IJCV_2004}. The source code is publicly accessible at http://yanweipang.com/papers. 

Intermediate results are also given to show the rationality of the assumptions mentioned before. 

\subsection{Results on the INRIA Pedestrian Database.}
In the INRIA dataset, the positive training set consists of 1208 normalized pedestrian windows, and the negative training set contains a mass of windows sampled from 1218 big and non-pedestrian images. The image size of the training window is $128 \times 64$ pixels, from which a 3780-dimensioned HOG feature vector is extracted. A linear SVM classifier $f( {\bf w})$ is obtained from the training sets.

As can be seen from Algorithms 2 and 3, the explicit parameters of iPW are $t_l $, $t_h $, $N_C^\ast $, $\alpha $ and $\gamma $. In our experiments, $t_l = - 2.0$ and $t_h = 0$ are used. Because rejection and acceptance particle windows are defined by not only $t_l $ and $t_h $, but also $r_R $ and $r_A $. So the parameters also include $r_R $ and $r_A $. In (8) and (10), the regions of rejection and acceptance are circular and isotropic whose size are determined by $r_R $ and $r_A$, respectively. However, because the height $h$ of the pedestrian is larger than its width $w$, it is more reasonable that the region of rejection and acceptance is rectangle. The size of the rectangle is represented as $r_R^x \times r_R^y $. Likewise, the size of region of acceptance is represented as $r_A^x \times r_A^y $. As stated in Section 4.2, $r_R^x $ and $r_R^y $ depend on the classifier response $f( {\bf w})$. In our experiments, $r_R^x $ and $r_R^y $ are quantized to 9 intervals according to the value of $f({\bf w})$. $r_R^x $ and $r_R^y $ also depend on the object width $w$ and height $h$ in question. Solid experiments are conducted to find rule for setting 
$r_R^x $ and $r_R^y $ according to $f({\bf w})$, $h$ and $w$. Table 2 shows how to choose $r_R^x $ and $r_R^y $. Note that $h = 128$ and $w = 64$. In Algorithm 3, we only use $r_R^x $ and $r_R^y $ when $f( {\bf w})$ belongs to the first four intervals. $r_A^x / w$ and $r_A^y / h$ are set to 0.16 and 0.16, respectively. $N_C^\ast = 0.5N_{iPW}$ is employed in the initialization step of Algorithm 3. The parameters 
$\alpha $ and $\gamma $ are set 0.2 and 0.7, respectively.

It is noted that regions of rejection and acceptance are cubic when scale factor is considered. The testing image is zoomed out by a factor $1 / 1.05$. If a window is rejected at current scale $s$, which belongs to the interval $N_{Interval} $, then the windows in adjacent scales ${s}'$ from $s\times 1.05^{3 - N_{Interval} }$ to $s / 1.05^{3 - N_{Interval} }$ with the size $0.8^\Delta r_R^x \times 0.8^\Delta r_R^y $ ($\Delta = \vert \log _{1.05}^{s / {s}'}  \vert$) are also rejected. If a window is accepted at current scale $s$, then the windows in adjacent scales ${s}'$ from $s\times 1.05^3$ to $s / 1.05^3$ with the size $0.8^\Delta r_A^x \times 0.8^\Delta r_A^y $ ($\Delta = \vert \log _{1.05}^{s / {s}'} \vert$) are also accepted.

\begin{table*}[!t]
\centering
\renewcommand{\arraystretch}{1.3}
\caption{Set $r_R^x $ and $r_R^y $ according to $f( {\bf w})$, $h$ and 
$w$.}
\begin{tabular}
{|c|c|c|c|c|c|c|c|c|c|}
\hline
$N_{Interval}$& 0& 1& 2& 3& 4& 5& 6& 7& 8 \\
\hline
$f(\bf{w})$& [-inf, -4.0]& [-4.0, -3.5]& [-3.5, -3.0]& [-3.0, -2.5]& [-2.5, -2.0]& [-2.0, -1.5]& [-1.5, -1.0]& [-1.0, -0.5]& [-0.5, 0.0] \\
\hline
$r_R^x / w$& 0.22& 0.18& 0.16& 0.12& 0.10& 0.06& 0.06& 0.02& 0.02 \\
\hline
$r_R^y / h$& 0.22& 0.18& 0.16& 0.12& 0.10& 0.06& 0.06& 0.02& 0.02 \\
\hline
\end{tabular}
\end{table*}

Table 3 compares iPW with MPW in terms of detection rate when they generate and examine the same number $N$ of particle windows. When $N = 2367$, the detection rate of iPW is 0.561 whereas the detection rate of MPW is 
0.469, meaning that the detection rate of iPW is 9.2{\%} higher than that of 
MPW. As $N$ decreases, the advantage of iPW becomes more remarkable.

\begin{table}[!t]
\centering
\renewcommand{\arraystretch}{1.3}
\caption{Detection rates vary with the number $N$ of particle windows when FPPI = 0.1.}
\begin{tabular}
{|c|c|c|c|c|c|c|}
\hline
$N$& 2367& 7100& 11833& 16567& 21300& 26033 \\
\hline
MPW& 0.469& 0.594& 0.614& 0.623& 0.625& 0.627 \\
\hline
iPW& 0.561& 0.620& 0.625& 0.627& 0.628& 0.630 \\
\hline
\end{tabular}
\end{table}

To further see the advantage of iPW, we show in Fig. 7 a specific detection 
result of iPW and MPW when $N$ is as small as 500. The red dots in Fig. 7(a) indicate the centers of 
particle windows in the last stage of MPW. Fig. 7(b) gives the final 
detection result by non-maximum suppression, where the man is detected whereas 
the woman is missed. The ambiguity particle windows of the last stage of iPW 
are shown in Fig. 7(c) and the final detection result is shown in Fig. 7(d). 
On the one hand, Fig. 7 demonstrates that MPW fails to detect the woman when a 
small number of particle windows is sampled whereas iPW is capable to 
localize both the woman and man. iPW generates the particle windows one by 
one. If the generated particle window has a lower classifier response, then the 
particle window (i.e., rejection particle window) will tell the next 
particle window not to sample from it and its neighboring region (i.e., region of 
rejection). As a result, other regions including the object regions will 
have larger possibility to be investigated. By contrast, MPW does not have 
the rejection mechanism. When the current particle windows do not contain 
clues of the objects, it is almost impossible for the latter particle 
windows to capture the location information of the objects.

\begin{figure}[!t]
\label{Fig7}
\centering
\subfigure[]{\label{Fig7:a}
\includegraphics[width=0.72in]{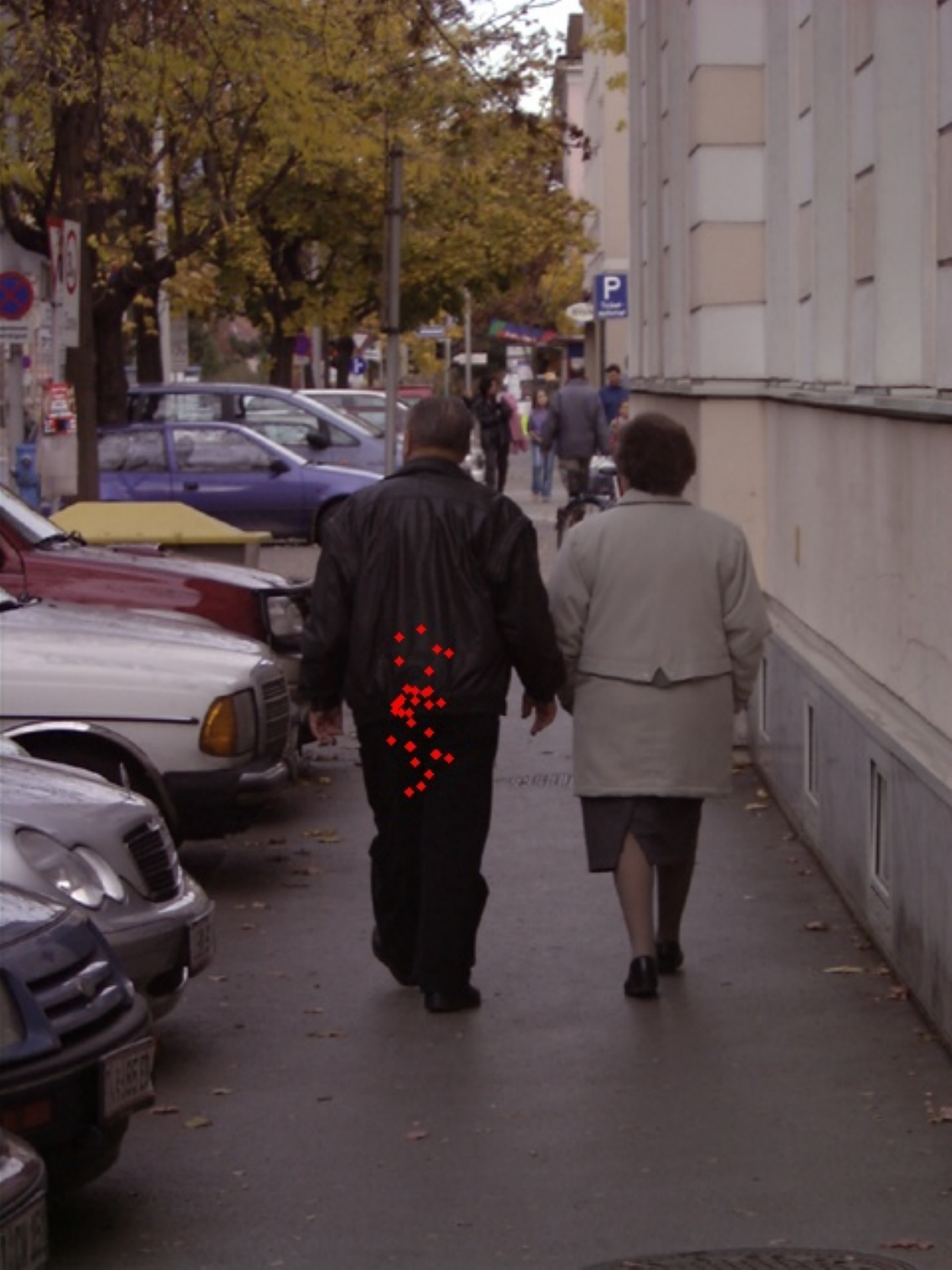}}
\hfil
\subfigure[]{\label{Fig7:b}
\includegraphics[width=0.72in]{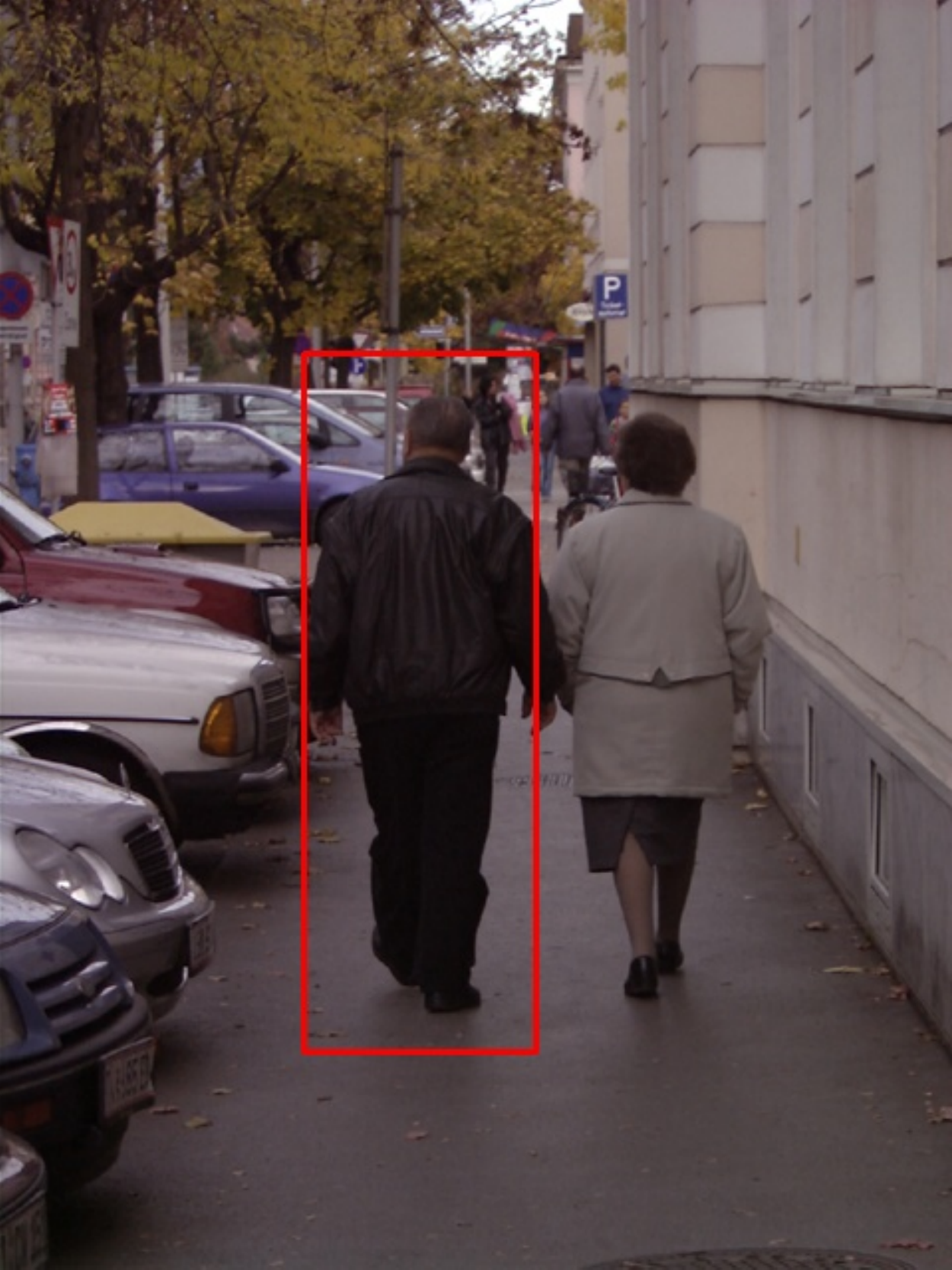}}
\hfil
\subfigure[]{\label{Fig7:c}
\includegraphics[width=0.72in]{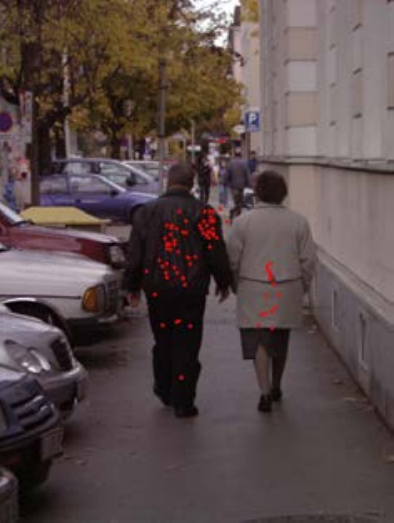}}
\hfil
\subfigure[]{\label{Fig7:d}
\includegraphics[width=0.72in]{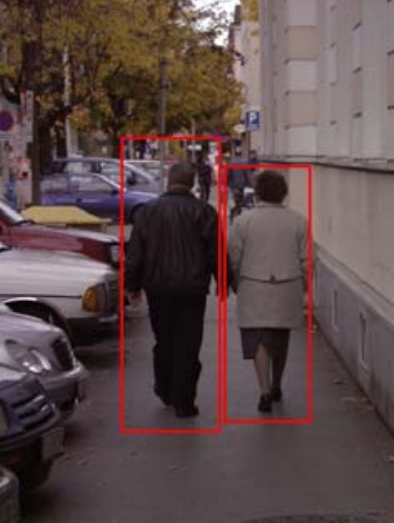}}
\caption{The detection results of iPW and MPW under 500 particle windows.
\subref{Fig7:a} The particle windows of the last stage in MPW.
\subref{Fig7:b} Final detection result of MPW.
\subref{Fig7:c} The particle windows of the last stage in iPW.
\subref{Fig7:d} Final detection result of iPW.}
\end{figure}

On the other hand, comparing Fig. 7(a) and (c), one can observe that MPW 
generates unnecessary too many particle windows around the man, whereas iPW 
can properly assign the limited number of particle windows to both the man 
region and the woman region. This phenomenon can be more clearly seen form 
Fig. 8. Fig. 8(a)-(c) give the updating process of particle windows in MPW, 
Fig. 8(d) shows that MPW is able to detect two persons if there is enough number of particle windows in the initialization. 
Fig. 8(e)-(g) give the updating process of particle windows in iPW, Fig. 8(h) 
shows that iPW detect even four pedestrians, including 
a false positive. 

%

\begin{figure}[!t]
\label{Fig8}
\centering
\includegraphics[width=3.2in]{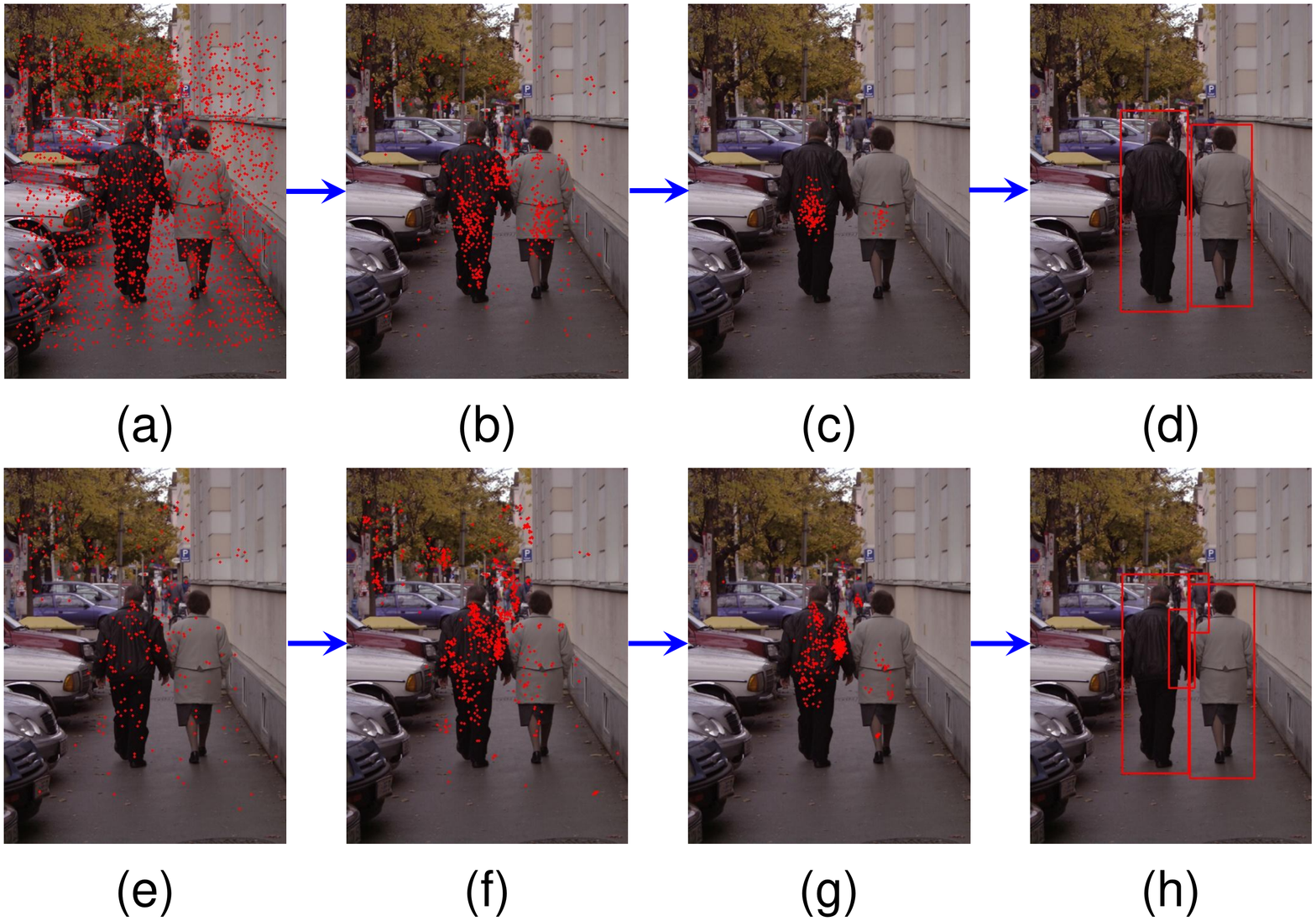}
\caption{ MPW generates unnecessary too many particle windows around the pedestrian regions. (a), (b) and (c) are the updating process of particle windows in MPW. (d) is the final detection result of MPW. (e), (f) and (g) are the updating process of particle windows in iPW. (h) is the final detection result of iPW.}
\end{figure}

Fig. 9 shows how $N_v^{iPW} (i) = \left| { {\bf W}_R } \right| + \left| {{\bf W}_A } \right|$ and $N_u^{iPW} (i) = N - \left| { {\bf W}_R } \right| - \left| { {\bf W}_A } \right|$ vary with the number $i$ of generated particle windows of iPW algorithm and how $N_v^{MPW} (i) = i$ and $N_u^{MPW} (i) = N - i$ vary with $i$ of MPW algorithm. One can see that the number $N_v^{iPW} (i)$ of visited windows of iPW grows much faster than that of MPW. Equivalently, 
the number $N_u^{iPW} (i)$ of unvisited windows of iPW drops much faster than that of MPW. So 
generating the same number of particle windows, iPW can classify (reject or accept) more windows (regions) than MPW. This explains why iPW obtains 
better detection accuracy than MPW when they use the same number of 
particle windows.

\begin{figure}[!t]
\label{Fig9}
\centering
\includegraphics[width=2.0in]{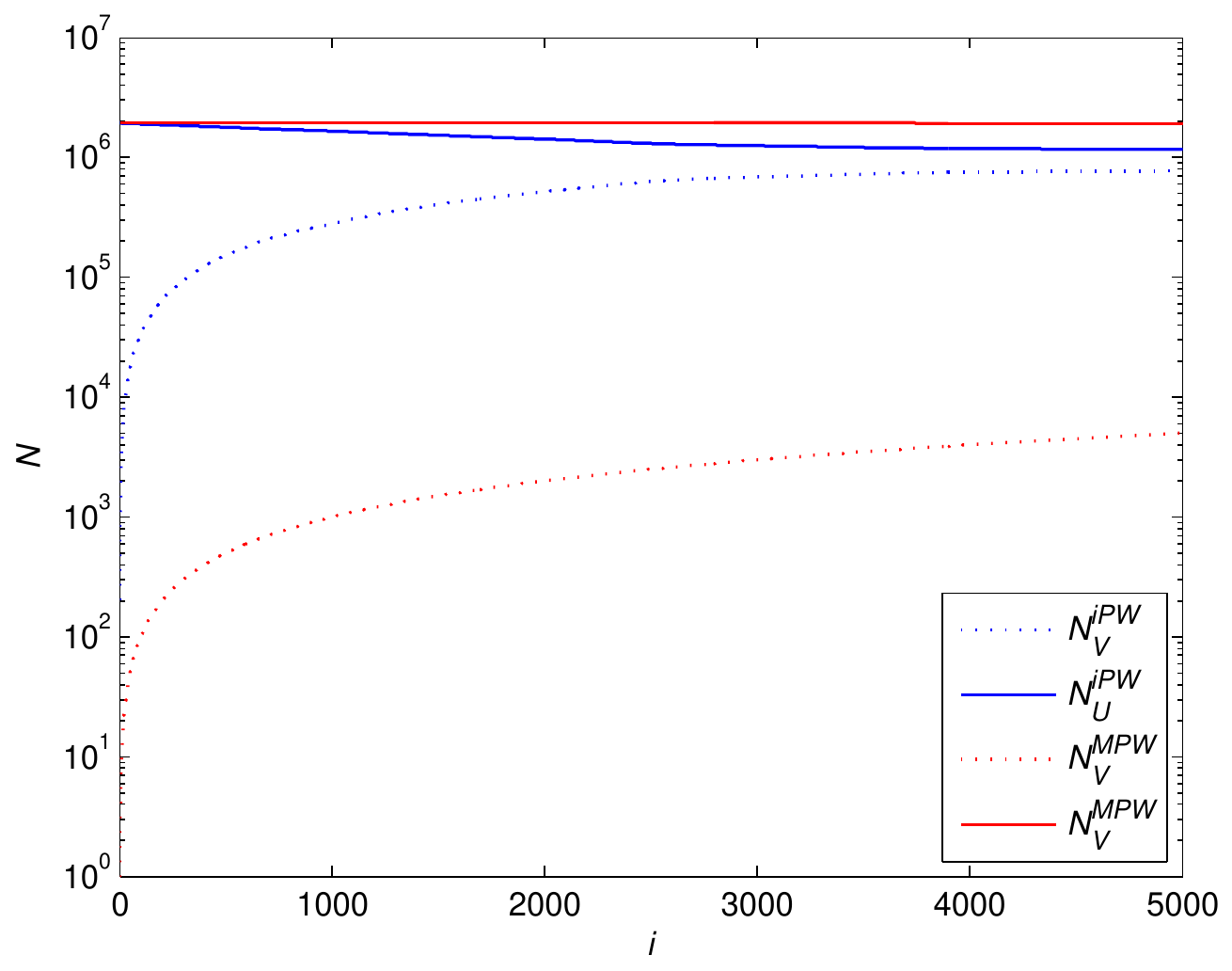}
\caption{The variations of $N_v^{MPW}$, $N_u^{MPW}$, $N_v^{iPW}$ and $N_u^{iPW}$.}
\end{figure}

If a smaller number of particle windows is generated, can iPW achieve the 
same detection accuracy as MPW? If it is true, then one can conclude that 
iPW is more efficient than MPW. To answer this question, SW is used as a baseline. It scans the image with the pixel stride 8 and 
scaling factor 1.05, and the number of scanned windows is denoted by $N_{SW} $. $N_{SW} $ varies with the size of testing image. For a $480 \times 640$ image, 
$N_{SW} $ is 47335. Let MPW generate $N_{MPW} = 0.3\times N_{SW} $ particle windows and iPW generate $N_{iPW} = 0.15\times N_{SW} $ particle windows. The resulting curves of miss rate vs. FFPI are plotted in Fig. 10. 
It is seen that these different window generation algorithms have very close 
operating points. For example, the miss rates of SW, MPW and iPW are 
23.4{\%}, 22.8{\%}, and 23.4{\%}, respectively when FFPI=1. At these 
operating points, the average values of $N_{SW} $, $N_{MPW} $ and $N_{iPW} $ in INRIA are shown in Table 4. Table 4 shows that to achieve the same 
operating point SW has to investigate 47335 windows whereas it is enough for 
iPW to generate and check 7099 windows. The detection time $T_{SW} $ of SW is 3.94 times of that (i.e., $T_{iPW} $) of iPW. Moreover, $N_{iPW} / N_{MPW} = 0.499$ means that using half of particle windows iPW can obtain the same detection 
rate as MPW. The ratio of detection time $T_{MPW} $ of MPW and detection time $T_{iPW} $ of iPW is 1.8, implying much higher efficiency of iPW than MPW.

\begin{figure}[!t]
\label{Fig10}
\centering
\includegraphics[width=2.5in]{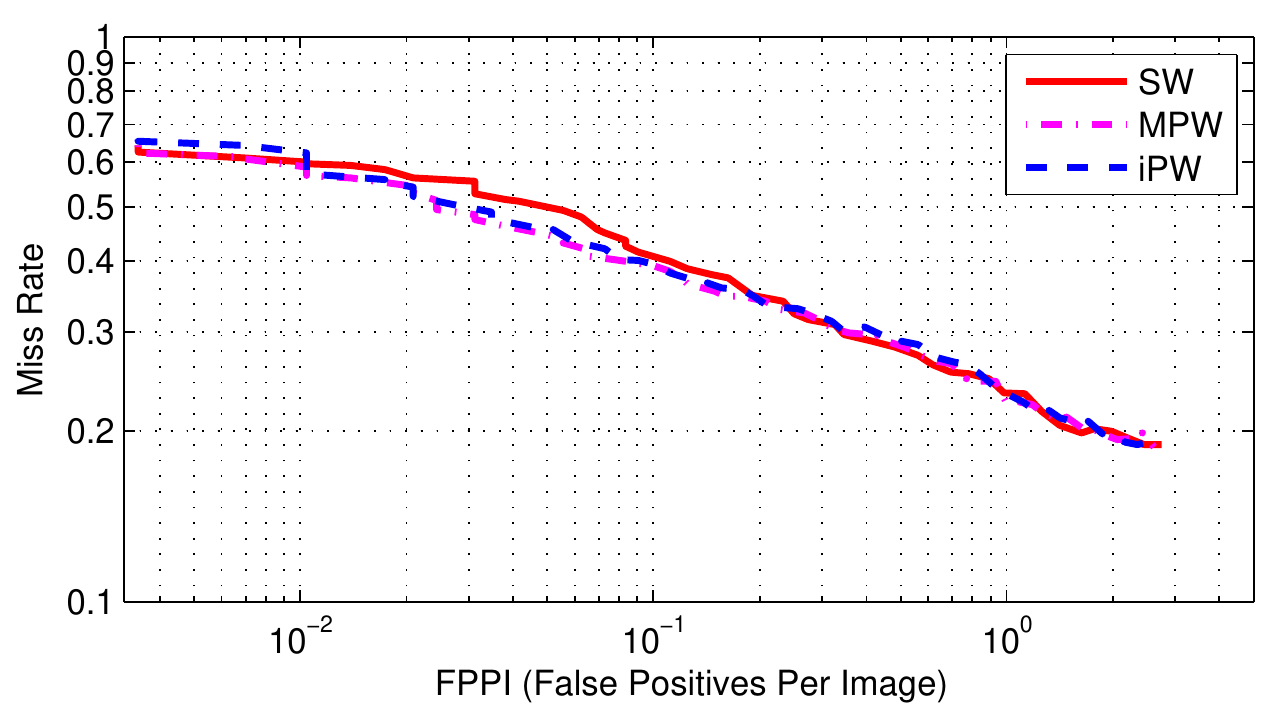}
\caption{DET curves comparing iPW with MPW and SW on INRIA.}
\end{figure}

\begin{table*}[!t]
\centering
\renewcommand{\arraystretch}{1.3}
\caption{Efficiency of SW, MPW and iPW on INRIA.}
\begin{tabular}
{|c|c|c|c|c|c|c|}
\hline
$N_{SW} $& $N_{MPW} $& $N_{iPW} $& $N_{iPW} / N_{MPW} $& $T_{MPW} / T_{iPW} $& $T_{SW} / T_{MPW} $& $T_{SW} / T_{iPW} $ \\
\hline
47335& 14200& 7099& 0.49& 1.80& 2.19& 3.94 \\
\hline
\end{tabular}
\end{table*}

\subsection{Results on the MIT-CMU Face Database.}
In Section 5.2, the feature and classifier are HOG and SVM, respectively. In 
this section, we evaluate iPW by using Haar-like features and cascade 
AdaBoost classifier for detecting faces in the standard MIT-CMU face 
database. The testing set consists of 125 images with 483 frontal 
faces. A 10 layers cascade model is learnt from 20000 normalized $20 \times 
20$ small face images and 5000 non-face large negative images.

Because the range of response of cascade AdaBoost is quite different from 
SVM, the low and high classifier thresholds $t_l $ and $t_h $ are also different from those in Section 5.2. Specifically, $t_l = 0.2$ and $t_h = 1.0$ are adopted. Consequently, the length $r_R $ and $r_A $ of regions of rejection and acceptance should be tuned. $r_R $ and $r_A $ are related to $f( {\bf w})$, $h$ and $w$. But the detection window is square, so $h = w$. The classifier response $f({\bf w})$ of a cascade AdaBoost is defined by $f({\bf w}) = j_ {\bf w} / 
L$, where $j_ {\bf w} $ is the index $j$ of the last stage which provides a positive classification for $ {\bf 
w}$, and $L = 10$ is total number of the stages of the cascade structure. The relationship 
between $r_R $, $f({\bf w}) $ and $h$ is given in Table 5 where $r_R / h$ monotonously decreases with $f({\bf w}) $. In Alogorithm 3, we only use the $r_R $ when $f( {\bf w})$ belongs to the first two values. $r_A^x / w$ and $r_A^y / h$ are set to 0.1 and 0.1, respectively. $N_C^\ast = 0.5N_{iPW}$ is employed in the initialization step of Algorithm 3. The parameters 
$\alpha $ and $\gamma $ are set 0.2 and 0.7, respectively.

\begin{table*}[!t]
\centering
\renewcommand{\arraystretch}{1.3}
\caption{Set $r_R $ according to $f({\bf w})$ and $h$.}
\begin{tabular}
{|c|c|c|c|c|c|c|c|c|c|c|}
\hline
$f({\bf w}) $& 0.0& 0.1& 0.2& 0.3& 0.4& 0.5& 0.6& 0.7& 0.8& 0.9 \\
\hline
$r_R / h$& 0.100& 0.090& 0.060& 0.050& 0.050& 0.040& 0.040& 0.030& 0.040& 0.030 \\
\hline
\end{tabular}
\end{table*}

Similar to Section 5.2, regions of rejection and acceptance are cubic. The 
testing image is zoomed out by a factor $1 / 1.15$. If a window is rejected at current scale $s$, the windows in neighboring scales ${s}'$ from $s\times 1.15$ to $s / 1.15$ with the size $0.5^\Delta r_R^x \times 0.5^\Delta r_R^y $ ($\Delta = \vert \log _{1.15}^{s / {s}'} \vert$) form the ${\bf W}_R $ of rejection particle windows. If a window is accepted at current scale $s$, then the windows in adjacent scales ${s}'$ from $s\times 1.15$ to $s / 1.15$ with the size $0.5^\Delta r_A^x \times 0.5^\Delta r_A^y $ ($\Delta = \vert \log _{1.15}^{s / {s}'} \vert$) are also accepted.

With the above parameters, iPW is applied to 125 testing images and 
detection rates corresponding to different number $N$ of particle windows are shown in Table 6. Table 6 also gives the detection 
rates of MPW. It is observed that iPW 
has almost higher detection rate in each case. When the number of particle 
windows is small, the advantage of iPW is more remarkable. For example, when 
$N = 25066$, the detection rate of iPW is 9.3{\%} higher than that of MPW.

\begin{table}[!t]
\centering
\renewcommand{\arraystretch}{1.3}
\caption{Detection rates vary with the number $N$ of particle windows when FPPI = 0.1.}
\begin{tabular}
{|c|c|c|c|c|c|c|}
\hline
$N$& 25066& 75200& 125333& 175466& 225600& 275733 \\
\hline
MPW& 0.583& 0.758& 0.788& 0.806& 0.811& 0.812 \\
\hline
iPW& 0.676& 0.792& 0.806& 0.813& 0.816& 0.819 \\
\hline
\end{tabular}
\end{table}

Fig. 11(b) and (d) respectively show the detection results of MPW and iPW 
when the number of particle windows is limited to 5000. Clearly, iPW is 
capable of detecting the two faces in the testing image whereas MPW does not 
detect any face at all. Fig. 11(a) and (c) show the centers of the particle 
windows generated in the last stage of MPW and iPW, respectively.

\begin{figure}[!t]
\label{Fig11}
\centering
\subfigure[]{\label{Fig11:a}
\includegraphics[width=0.72in]{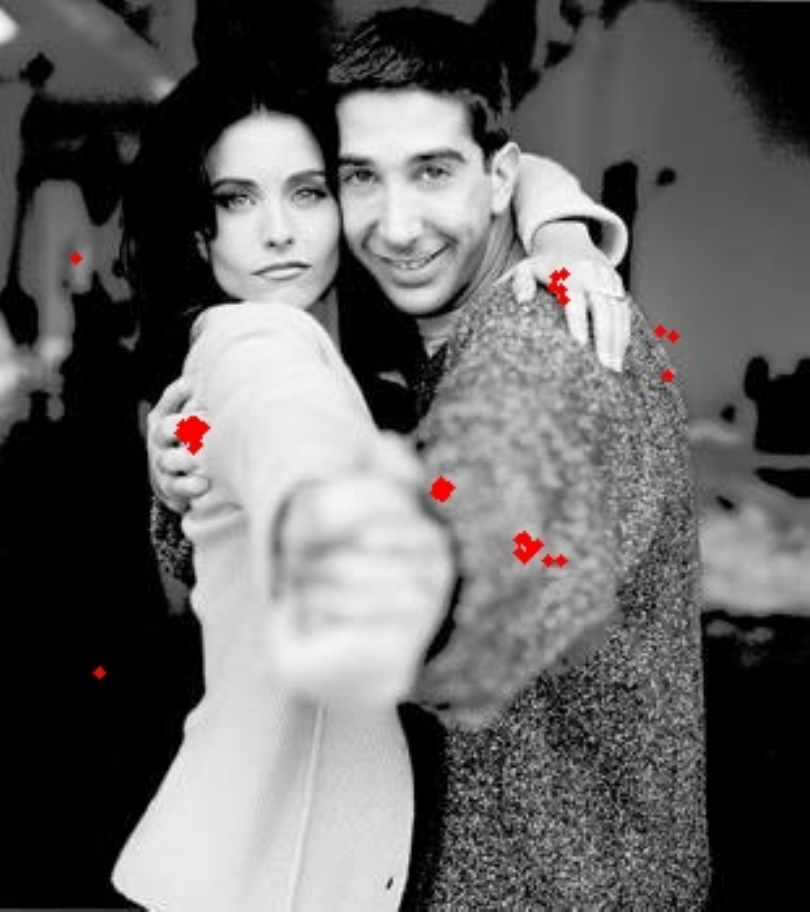}}
\hfil
\subfigure[]{\label{Fig11:b}
\includegraphics[width=0.72in]{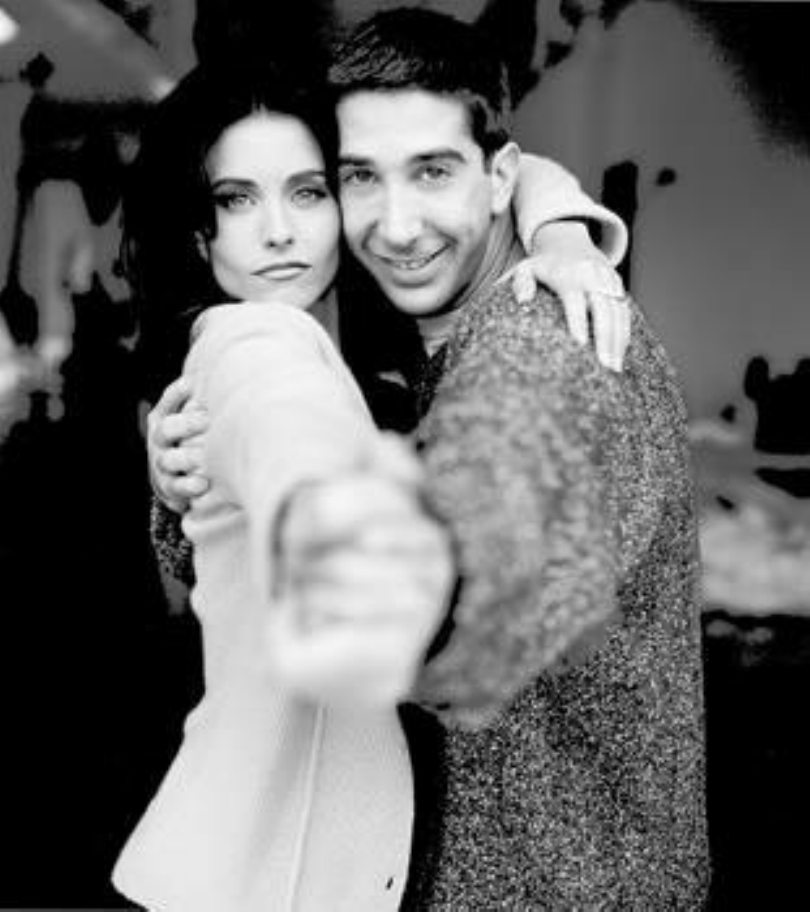}}
\hfil
\subfigure[]{\label{Fig11:c}
\includegraphics[width=0.72in]{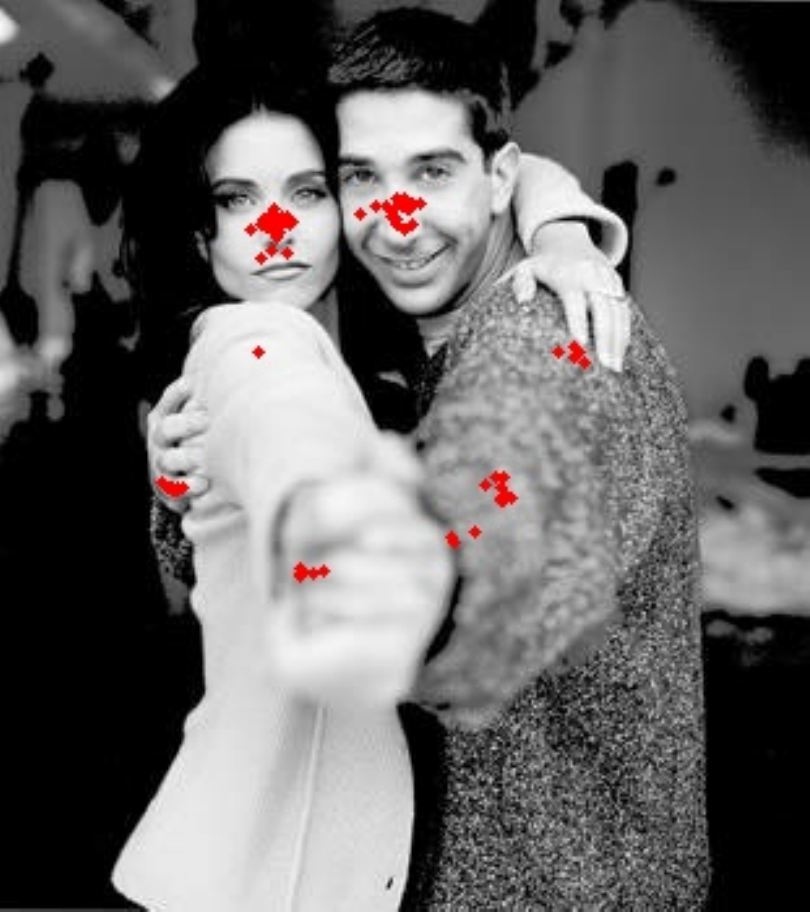}}
\hfil
\subfigure[]{\label{Fig11:d}
\includegraphics[width=0.72in]{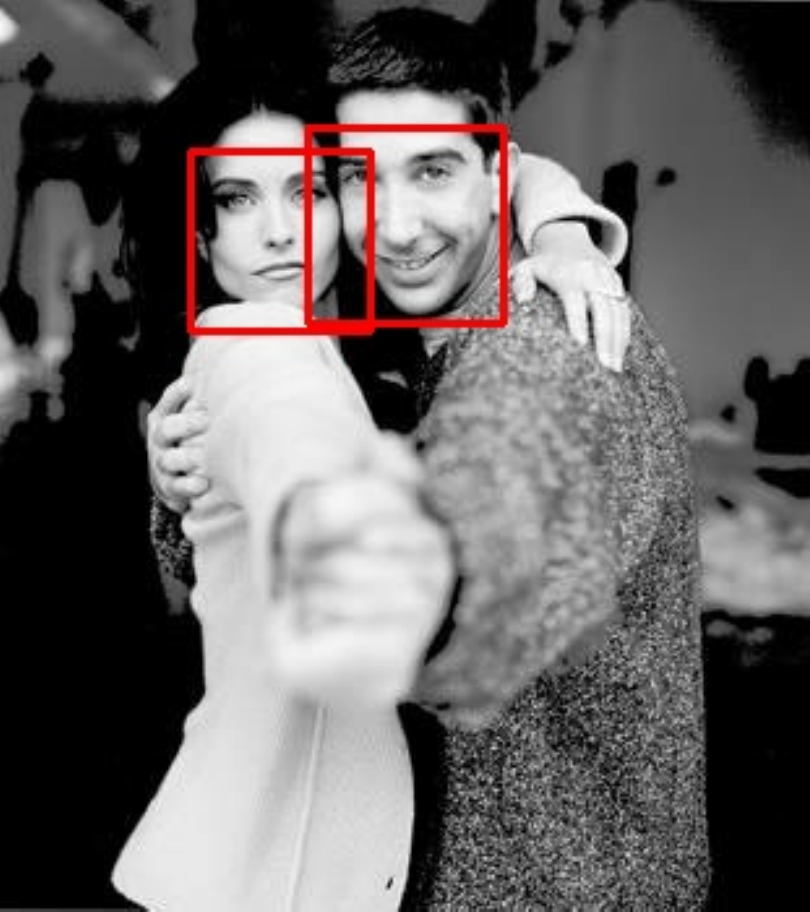}}
\caption{The detection results of iPW and MPW under 5000 particle windows.
\subref{Fig11:a} The particle windows of the last stage in MPW.
\subref{Fig11:b} Final detection result of MPW.
\subref{Fig11:c} The particle windows of the last stage in iPW.
\subref{Fig11:d} Final detection result of iPW.}
\end{figure}

Fig. 12(a)-(d) show that when the number of particle windows of MPW is upto 
20000, MPW can detect the two faces in the testing image. But MPW assigns too 
many particle windows around object and object-like regions. The iterations 
of iPW are shown in Fig. 12(e)-(h) which assign proper number of particle 
windows around object regions. The intuition is that if there are a few 
acceptance particle windows around the object, then it is no longer 
necessary to generate additional particle windows around the object. 
Instead, the opportunity should be given to check other region.
%
%

\begin{figure}[!t]
\label{Fig12}
\centering
\centering
\includegraphics[width=3.2in]{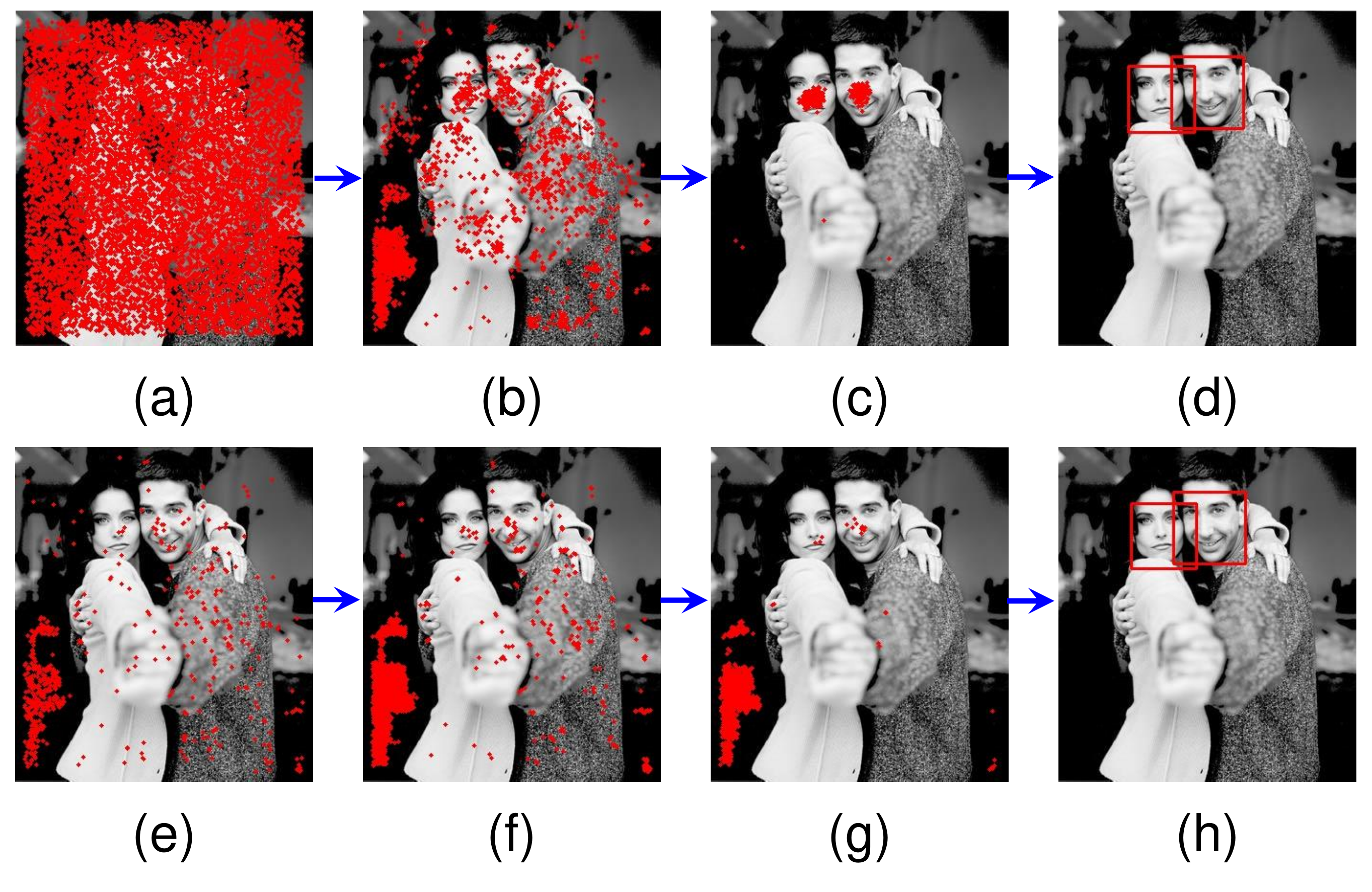}

\caption{ (a), (b) and (c) are the updating process of particle windows in MPW. (d) is the final detection result of MPW. (e), (f) and (g) are the updating process of particle windows in iPW. (h) is the final detection result of iPW.}
\end{figure}

Finally, experiments are conducted to see whether iPW can achieve comparable 
face detection results as MPW if a smaller number of particle windows 
is used. As in pedestrian detection experiments, SW is also used as 
baseline. It slides the testing image with pixel stride 2 and scale factor 
1.25. As a result, SW generates 621816 and 3941097 windows for $454 \times 
628$ and $1024 \times 1280$ images, respectively. Limit the numbers of 
particle windows in MPW and iPW to $N_{MPW} = 0.25\times N_{SW} $ and $N_{iPW} = 0.13\times N_{SW} $, respectively. The resulting ROC curves of SW, MPW and iPW are shown in 
Fig. 13. It is seen from Fig. 13 that SW is almost consistently inferior to both 
MPW and iPW. Even only 0.13 fraction of windows are used, iPW can obtain 
higher detection rates than SW. Moreover, one can see Fig. 13 that the miss 
rates of iPW are almost identical to that of MPW.

\begin{figure}[!t]
\label{Fig13}
\centering
\includegraphics[width=2.5in]{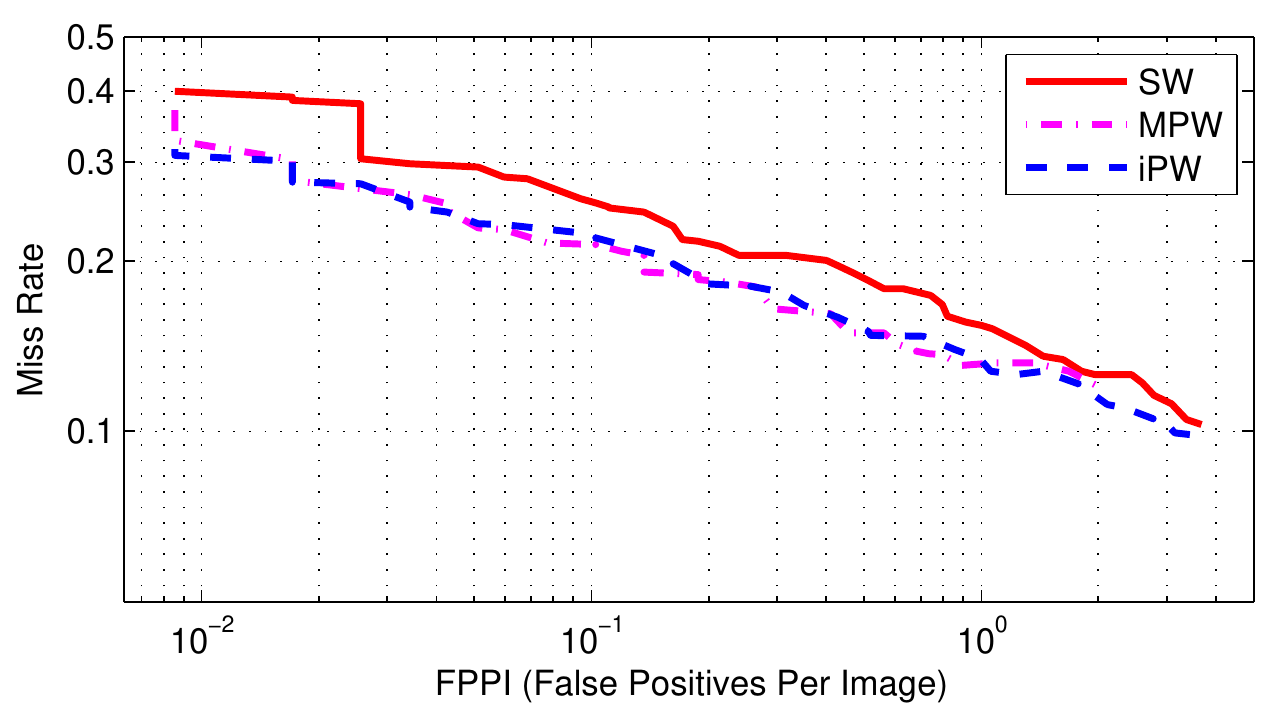}
\caption{DET curves comparing iPW with MPW and SW on MIT-CMU.}
\end{figure}

\begin{table*}[!t]
\centering
\renewcommand{\arraystretch}{1.3}
\caption{Efficiency of SW, MPW and iPW on MIT-CMU.}
\begin{tabular}
{|c|c|c|c|c|c|c|}
\hline
$N_{SW}$ & $N_{MPW} $& $N_{iPW} $& $N_{iPW} / N_{MPW} $& $T_{MPW} / T_{iPW} $& $T_{SW} / T_{MPW} $& $T_{SW} / T_{iPW} $ \\
\hline
501334& 125333& 65173& 0.520& 1.50& 1.15& 1.73 \\
\hline
\end{tabular}
\end{table*}

Table 7 shows the testing time $T_{SW} $ of SW is 1.15 times and 1.73 times of MPW and iPW, respectively, when the 
number of generated windows of SW, MPW and iPW are 501334, 125333 and 65173. 
The miss rates of the three algorithms correspond to the point in Fig. 13 
with FPPI=1. The speedup effect in face detection is not as significant as 
pedestrian detection. The reason is that the number of HOG features is fixed 
in pedestrian detection but the number of Haar-like features varies with the 
response of cascade classifier.

\section{Conclusion}
In this paper, we have proposed how to improve MPW. The proposal distribution of MPW mainly relies on 
the regions of support. In contrast, the proposed iPW and siPW algorithms 
construct the proposal distribution based on the proposed concepts of 
rejection, acceptance, and ambiguity particle windows which are defined by low 
and high thresholds of the classifier response. Both the rejection and 
acceptance particle windows are used for reducing the search space. The 
existence of the objects is reflected in the acceptance particle windows and 
the main clue of object locations is contained in ambiguity particle 
windows. Specifically, our proposal distribution is a weighted average of 
dented uniform distribution and dented Gaussian distribution which are 
dented by the rejection and acceptance particle windows. An important 
characteristic of the proposed algorithms is that single particle windows is 
generated in each stage, which makes iPW to run in an incremental manner. 
Experimental results have shown that iPW is about two times efficient than 
MPW.


%

\appendices
\end{document}